%% file: arXiv.tex
\documentclass[manuscript,screen,nonacm]{acmart}
\renewcommand\footnotetextcopyrightpermission[1]{}
\AtBeginDocument{%
  }

\setcopyright{acmlicensed}
\copyrightyear{2018}
\acmYear{2018}
\acmDOI{XXXXXXX.XXXXXXX}

\acmJournal{CSUR}
\acmVolume{37}
\acmNumber{4}
\acmArticle{111}
\acmMonth{8}

\usepackage{xcolor}
\usepackage{array}
\usepackage{multirow, makecell}
\usepackage{fontawesome5}
\usepackage{tabularx}
\usepackage{ulem}
\usepackage{dsfont}
\usepackage{subcaption}
\usepackage{mathtools}
\usepackage{longtable}
\usepackage{ltxtable}
\usepackage{hyperref}

\usepackage{tabularx}
\usepackage[table]{xcolor}
\usepackage{booktabs}

\usepackage{svg}
\usepackage{graphicx}
\usepackage{eso-pic}

\begin{document}

\AddToShipoutPictureFG*{%
  \AtPageUpperLeft{%
    \raisebox{-12.0em}{
      \hspace{7.5em}
      \includegraphics[height=1.8em]{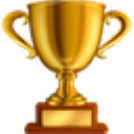}
    }%
  }%
}

\input{arXiv_main}

\bibliographystyle{ACM-Reference-Format}
\bibliography{ref.bib}

\clearpage
\appendix

\begin{center}
    {\LARGE \textbf{SUCCESS-GS}: Survey of Compactness and Compression for \\Efficient Static and Dynamic Gaussian Splatting}
\end{center}

\begin{center}
    {\Large Supplementary Material}
\end{center}

\input{arXiv_supplement}

\end{document}

%% file: arXiv_main.tex
\title{\hspace*{1.5em}SUCCESS-GS: Survey of Compactness and Compression for Efficient Static and Dynamic Gaussian Splatting}

\author{Seokhyun Youn}
\email{hisn16@cau.ac.kr}
\authornote{These authors contributed equally to this paper.}
\affiliation{%
  \institution{Chung-Ang University}
  \country{South Korea}
}

\author{Soohyun Lee}
\authornotemark[1]
\email{nata1225@khu.ac.kr}
\affiliation{%
  \institution{Kyung Hee University}
  \country{South Korea}
}

\author{Geonho Kim}
\authornotemark[1]
\email{joelkimgh@cau.ac.kr}
\affiliation{%
  \institution{Chung-Ang University}
  \country{South Korea}
}

\author{Weeyoung Kwon}
\email{weeyoungkwon@cau.ac.kr}
\affiliation{%
  \institution{Chung-Ang University}
  \country{South Korea}
}

\author{Sung-Ho Bae}
\authornote{Co-corresponding authors.}
\affiliation{%
   \institution{Kyung Hee University}
  \city{Seoul}
  \country{South Korea}}
\email{shbae@khu.ac.kr}
\orcid{000-0002-3389-1159}

\author{Jihyong Oh}
\authornotemark[2]
\affiliation{%
   \institution{Chung-Ang University}
  \city{Seoul}
  \country{South Korea}}
\email{jihyongoh@cau.ac.kr}
\orcid{0000-0002-1627-0529}

\renewcommand{\shortauthors}{Youn \textit{et al}.}

\begin{abstract}
\small\url{https://cmlab-korea.github.io/Awesome-Efficient-GS/}

\vspace{0.8em}
 3D Gaussian Splatting (3DGS) has emerged as a powerful explicit representation enabling real-time, high-fidelity 3D reconstruction and novel view synthesis. However, its practical use is hindered by the massive memory and computational demands required to store and render millions of Gaussians. These challenges become even more severe in 4D dynamic scenes. To address these issues, the field of Efficient Gaussian Splatting has rapidly evolved, proposing methods that reduce redundancy while preserving reconstruction quality. This survey provides the first unified overview of efficient 3D and 4D Gaussian Splatting techniques. For both 3D and 4D settings, we systematically categorize existing methods into two major directions, Parameter Compression and Restructuring Compression, and comprehensively summarize the core ideas and methodological trends within each category. We further cover widely used datasets, evaluation metrics, and representative benchmark comparisons. Finally, we discuss current limitations and outline promising research directions toward scalable, compact, and real-time Gaussian Splatting for both static and dynamic 3D scene representation.
\end{abstract}

\maketitle
\input{sections/Sec.1.introduction}
\input{sections/Sec.2.preliminary}

\input{sections/Sec.3.static}
\input{sections/Sec.3.dynamic}
\input{sections/Sec.4.dataset}
\input{sections/Sec.6.limitations_and_future_directions}
\input{sections/Sec.7.conclusion}

%% file: sections/Sec.1.introduction.tex
\section{INTRODUCTION}
\textcolor{black}{3D scene representation and photorealistic novel view synthesis are crucial for a wide range of applications, including virtual reality (VR)~\cite{b133,b134}, augmented reality (AR)~\cite{b135}, media generation~\cite{b138}, autonomous driving~\cite{b139}, and large-scale visualization~\cite{b140,b141}. The evolution of these tasks has been dramatically accelerated by the emergence of Neural Radiance Fields (NeRF)~\cite{b32}. By directly training \textit{implicit} neural networks through differentiable volume rendering, NeRF has achieved high-quality 3D reconstruction from 2D images. However, despite its impressive visual quality, NeRF suffers from extremely slow training rendering speed, since it requires a large number of ray samples per pixel to generate a single image.~\cite{b144,b145,b146,b147,b148,b149,b150}}

\textcolor{black}{To overcome the limitations of such implicit representations, 3D Gaussian Splatting (3DGS)~\cite{b1} revisits the idea of \textit{explicit} 3D scene representation. Similar to traditional 3D representations such as point clouds~\cite{b151,b152,b153} and meshes~\cite{b154,b155}, 3DGS explicitly represents a scene using a collection of primitives. Unlike these classical forms, however, 3DGS models the scene as a set of anisotropic 3D Gaussians, each parameterized by a set of learnable attributes including position, scale, rotation, opacity, and view-dependent color coefficients (see Sec.~\ref{Sec. 2.1} for more details). By learning these attributes directly from multi-view images, 3DGS not only achieves high-quality scene reconstruction, but also enables real-time rendering performance, which is infeasible for NeRF-based approaches.}

\textcolor{black}{Further, recent methods can reconstruct a 4D scene representation from monocular or multi-view video inputs. Dynamic 3D/4D Gaussian Splatting extends conventional static 3DGS into the temporal dimension, enabling the reconstruction of not only static scenes and objects but also dynamic ones exhibiting non-rigid motion. By modeling spatial and temporal variations within a unified spatio-temporal Gaussian framework, 4DGS provides continuous motion representation and temporally coherent rendering (see Sec.~\ref{Sec. 2.2} for more details). This capability opens up new possibilities for dynamic scene editing~\cite{b142}, motion capture~\cite{b97}, and realistic simulation in robotics and autonomous driving~\cite{b143}, where time-varying geometry and appearance are crucial.}

\textcolor{black}{However, despite these advantages, the practical deployment of Gaussian Splatting–based methods remains challenging due to their massive memory footprint and computational overhead~\cite{b67}. A typical high-resolution static scene often contains millions of 3D Gaussians, which require significantly more memory than NeRF-based models~\cite{b4}. Furthermore, 4D scene reconstruction with Gaussian Splatting demands even greater memory consumption, as each Gaussian must encode temporal information across multiple frames~\cite{b96}.}

\textcolor{black}{To overcome these bottlenecks, a new research direction has emerged under the umbrella of Efficient Gaussian Splatting, aiming to reduce the redundancy of Gaussian primitives while preserving rendering quality. As various studies have been conducted, research on Efficient Gaussian Splatting can be categorized into the following groups:
\begin{itemize}
    \item \textbf{Parameter Compression (Sec.~\ref{Sec. 3.1}, Sec.~\ref{Sec. 4.1})}: This approach directly compresses the attributes of Gaussians using techniques such as pruning, quantization, and entropy coding.
    \item \textbf{Restructuring Compression (Sec.~\ref{Sec. 3.2}, Sec.~\ref{Sec. 4.2})}: This approach performs compression by redesigning the structure of the original Gaussian Splatting framework through methods such as hierarchical anchors, neural integration, or geometry-aware organization.
\end{itemize}}

\textcolor{black}{This survey provides the first comprehensive overview of Efficient 3D and 4D Gaussian Splatting (Sec.~\ref{static}, Sec.~\ref{dynamic}).
For both static 3DGS and dynamic 3D/4D Gaussian Splatting, we categorize existing methods into Parameter Compression (Sec.~\ref{Sec. 3.1}, Sec.~\ref{Sec. 4.1}) and Restructuring Compression (Sec.~\ref{Sec. 3.2}, Sec.~\ref{Sec. 4.2}), further organizing them into subcategories. For each class, we summarize the core methodology, mathematical formulation, and representative works. In addition, this survey provides a systematic overview of widely used datasets and evaluation metrics in Gaussian Splatting research, and presents comparisons of recent methods to establish a fair and consistent basis for performance evaluation (Sec.~\ref{DatasetsAndEvaluation}). Finally, we comprehensively discuss the limitations of existing approaches and propose key future directions that may guide the development of more efficient Gaussian Splatting techniques (Sec.~\ref{Sec. 6}).
By establishing a unified taxonomy and connecting isolated research threads, this survey aims to serve as a foundation for the further development of scalable, efficient, and robust Gaussian Splatting frameworks for static and dynamic 3D scene representation. 
}

%% file: sections/Sec.2.preliminary.tex
\section{PRELIMINARY}
\label{Sec. 2}
\textcolor{black}{
All mathematical notations used in this survey are summarized in the supplementary material.}
\subsection{Static 3D Gaussian Splatting (3DGS)}
\label{Sec. 2.1}
 
3DGS represents a 3D scene as a set of anisotropic 3D Gaussians and directly projects them into 2D screen space for rendering. Unlike Implicit Neural Representations (e.g., \textcolor{black}{NeRF~\cite{b32}}), 3DGS can be optimized without neural networks and supports real-time rendering through a CUDA-based tile rasterization pipeline. This section describes the initialization, rendering pipeline, and optimization process of 3DGS. An overall overview of the 3DGS framework is shown in Fig.~\ref{figure1}.
\begin{figure}[t]
    \centering
    \includegraphics[width=\columnwidth]{acmart-primary/acmart-primary/Figure1_v2.pdf}
     \caption{\textbf{Overview of the static 3D Gaussian Splatting (3DGS) pipeline.} A scene is represented as a set of 3D Gaussians with attributes including position $\boldsymbol{\mu}$, scale $\boldsymbol{s}$, rotation $\boldsymbol{r}$, spherical harmonics (SH) coefficients $\boldsymbol{h}$, and opacity $\sigma$ (Sec.~\ref{Sec. 2.1.1}). Each Gaussian is projected into the camera coordinate system and rendered onto the image plane through a differentiable tile rasterizer (Sec.~\ref{Sec. 2.1.3}, Sec.~\ref{Sec. 2.1.4}). The rendering is optimized by minimizing the difference between predicted and GT images (Sec.~\ref{Sec. 2.1.5}). To improve reconstruction quality, adaptive density control clones or splits Gaussians based on view-space gradient signals while pruning nearly transparent Gaussians (Sec.~\ref{Sec. 2.1.6}). In addition, efficiency-oriented strategies such as pruning, quantization, and entropy coding are applied to compress Gaussian attributes (Sec.~\ref{Sec. 3.1}).}
    \label{figure1}
\end{figure}

\subsubsection{Initialization and Attribute Definition} \label{Sec. 2.1.1}

3DGS requires multi-view RGB images of the same scene along with their corresponding camera pose information as input. Therefore, before the actual initialization of the 3D Gaussians, a Structure-from-Motion (SfM) tool (e.g., COLMAP\textcolor{black}{~\cite{b31}}) is used to initialize the sparse point cloud of the scene and recover the camera poses for each image. The sparse point cloud is then used for Gaussian initialization, where each point is initialized as a single 3D Gaussian. The estimated camera poses are used later in the projection stage.

Each Gaussian has the following attributes:

\begin{itemize}
    \item \textbf{Position} $\boldsymbol{\mu} \in \mathbb{R}^3$: \textcolor{black}{It represents} the center coordinates of the 3D Gaussian. This attribute is initialized using the world coordinates from the sparse point cloud.
    \item \textbf{Scale} $\boldsymbol{s} \in \mathbb{R}^3$: \textcolor{black}{It indicates} the spatial extent occupied by the 3D Gaussian.
    \item \textbf{Rotation} $\boldsymbol{r} \in \mathbb{R}^4$: \textcolor{black}{It represents} the orientation of the 3D Gaussian, initialized as a unit quaternion.
    \item \textbf{View-dependent Spherical Harmonics (SH) coefficients} \textcolor{black}{$\boldsymbol{h} \in \mathbb{R}^{(d+1)^2\times3}$}: \textcolor{black}{They represent} colors that vary with the viewing direction. \textcolor{black}{$d$ is a parameter that denotes the degree of the SH, which determines the level of detail in representing view-dependent colors in the scene. It is commonly set to $d=3$ in practice.} Since the spherical harmonics functions correspond to each RGB channel, the SH coefficients are also defined accordingly.
    \item \textbf{Opacity} ${\sigma} \in \mathbb{R}$: Each 3D Gaussian has an opacity value in the range $[0,1]$, stored with \textcolor{black}{$M$}-bit precision. As the opacity value approaches 0, the 3D Gaussian becomes more transparent, while as it approaches 1, it becomes more opaque. \textcolor{black}{In the original 3DGS paper~\cite{b1}, it uses $M=8$ to store Gaussian opacity.}
\end{itemize}

Based on the above, in this survey, the set of $N$ 3D Gaussians \textcolor{black}{$\mathcal{G}$ at a specific camera \textcolor{black}{pose} and their corresponding attributes are denoted as follows:
\begin{equation}\label{eqn:(1)}
\mathcal{G} = \{ G_i = (\boldsymbol{\mu}_i, \boldsymbol{s}_i, \boldsymbol{r}_i, \boldsymbol{h}_i, \sigma_i) \}_{i=1}^N  \ .
\end{equation}}

Let $G_i$ denote the $i$-th Gaussian among $N$ Gaussians, with position $(\boldsymbol{\mu}_i)$, scale $(\boldsymbol{s}_i)$, rotation $(\boldsymbol{r}_i)$, SH coefficients $(\boldsymbol{h}_i)$, and opacity \textcolor{black}{($\sigma_i$)}. Each attribute of the Gaussian is optimized through the training process.

\subsubsection{Mathematical Definition of a 3D Gaussian} \label{Sec. 2.1.2}
When the center vector of the $i$-th Gaussian in \textcolor{black}{coordinate system} is $\boldsymbol{\mu}_i$, the Gaussian is defined as
\begin{equation}\label{eqn:(2)}
G_i(\boldsymbol{x}) = \exp( -\frac{1}{2} (\boldsymbol{x} - \boldsymbol{\mu}_i)^\top \boldsymbol{\Sigma}_i^{-1} (\boldsymbol{x} - \boldsymbol{\mu}_i)),
\end{equation}
where $\boldsymbol{x}$ is an arbitrary point in \textcolor{black}{the world coordinate} and $\boldsymbol{\Sigma}_i$ denotes the covariance matrix of the $i$-th Gaussian. The covariance matrix is formulated as
\begin{equation}\label{eqn:(3)}
\boldsymbol{\Sigma}_i = \boldsymbol{R}_i \boldsymbol{S}_i \boldsymbol{S}_i^\top \boldsymbol{R}_i^\top,
\end{equation}
where $\boldsymbol{R}_i$ and $\boldsymbol{S}_i$ are the rotation and scaling matrices of the Gaussian, respectively, which are derived from the Gaussian’s attributes $\boldsymbol{r}_i$ and $\boldsymbol{s}_i$.

\subsubsection{Projection} \label{Sec. 2.1.3}

Let $\boldsymbol{W}$ be the coordinate transformation from the world coordinates to the camera coordinates, and $\boldsymbol{J}$ be the Jacobian of the camera coordinate to screen coordinate projection. The covariance matrix is formulated as:
\begin{equation}\label{eqn:(4)}
\boldsymbol{\Sigma}_i' = \boldsymbol{J}\boldsymbol{W} \boldsymbol{\Sigma}_i \boldsymbol{W}^\top \boldsymbol{J}^\top.
\end{equation}

This results in the representation of each 3D anisotropic Gaussian as a 2D ellipse in the screen coordinate system.

\subsubsection{Differentiable Tile Rasterization} \label{Sec. 2.1.4}

The rendering pipeline proceeds as follows. First, the image is divided into \textcolor{black}{$T \times T$} pixel tiles. \textcolor{black}{According to the original 3DGS paper, the tile size $T$ was set to 16.} Next, for each projected Gaussian, the overlapping tiles are determined. Finally, per-pixel rasterization is performed to compute the contribution of each Gaussian to the pixels within the overlapping tiles.

For each Gaussian $G_i$, the alpha value at pixel $\boldsymbol{p}$ is given by
\begin{equation}\label{eqn:(5)}
\alpha_i(\boldsymbol{p}) = \sigma_i \cdot g_i(\boldsymbol{p}) \ ,
\end{equation}

where $g_i(\boldsymbol{p})$ denotes the value of the projected 2D Gaussian at pixel $\boldsymbol{p}$, computed as
\begin{equation}\label{eqn:(6)}
g_i(\boldsymbol{p}) = \exp\left( -\frac{1}{2}(\boldsymbol{p} - \boldsymbol{\mu}'_i)^\top (\boldsymbol{\Sigma}'_i)^{-1} (\boldsymbol{p} - \boldsymbol{\mu}'_i) \right).
\end{equation}

Here, $\boldsymbol{\mu}'_i$ denotes the 2D projected center of the $i$-th Gaussian after applying the camera projection, and $\boldsymbol{\Sigma}'_i$ is the 2D covariance matrix obtained by projecting the 3D covariance $\boldsymbol{\Sigma}_i$ using the Jacobian of the projection function.

The final color of pixel $\boldsymbol{p}$ is obtained via alpha blending, which combines the color $\boldsymbol{c}_i$ predicted from the Gaussian’s SH coefficients with the corresponding depth information:
\begin{equation}\label{eqn:(7)}
C(\boldsymbol{p}) = \sum_{i \in \mathcal{N}} \boldsymbol{c}_i \alpha_i(\boldsymbol{p}) \prod_{j=1}^{i-1} \left( 1 - \alpha_j(\boldsymbol{p}) \right).
\end{equation}

In this expression, \textcolor{black}{$\mathcal{N}$} denotes the total number of Gaussians projected onto the pixel. \textcolor{black}{$\boldsymbol{c}_i$ is the view-dependent base color of the Gaussian, computed from its SH coefficients and the camera viewing direction.} The rasterizer is fully differentiable, allowing gradients to be backpropagated to all Gaussian parameters.

\subsubsection{Optimization} \label{Sec. 2.1.5}

The reconstruction is optimized by minimizing the difference between the predicted \textcolor{black}{2D} image $I_{pred}$ and the ground truth \textcolor{black}{(GT)} $I_{gt}$:
\begin{equation}\label{eqn:(8)}
L = (1 - \lambda) \cdot L_1(I_{pred}, I_{gt}) + \lambda \cdot ({1 - SSIM(I_{pred}, I_{gt}))} \ ,
\end{equation}
where $L_1(\cdot)$ is the \textcolor{black}{pixel-wise mean absolute error}, and $SSIM$\textcolor{black}{~\cite{b33}} is the structural similarity index. \textcolor{black}{$\lambda$ is a weighting factor that controls the relative contribution of the $L_1(\cdot)$ and the $SSIM$ loss in the total loss function.}

\subsubsection{Adaptive Density Control}\label{adaptive-density-control} \label{Sec. 2.1.6}

Since the initial Gaussians are generated from a sparse point cloud obtained via SfM, they are insufficient to represent the entire scene. Therefore, during training, the number of Gaussians is adaptively increased by duplicating or splitting existing Gaussians. \textcolor{black}{ To select Gaussians for densification, the view-space position gradients of the Gaussians are considered. A large gradient indicates a high reconstruction error in the corresponding region, implying that more Gaussians are required to represent the scene in greater detail. Specifically, if the view-space position gradients of a Gaussian exceed a certain threshold $\tau_{pos}$, which is set to 0.002 in the 3DGS paper, densification is performed. In this process, Gaussians with relatively small scales $\boldsymbol{s}$ are cloned, while Gaussians with larger scales are split into smaller Gaussians and distributed around the original position. This densification step is performed at fixed epoch intervals, which reduces unnecessary computational cost and ensures efficient training.} In addition, if the opacity $\sigma_i$ of a Gaussian is less than a certain threshold $\epsilon_{\sigma}$ , it is considered nearly transparent and thus negligible for scene reconstruction, and such Gaussians are pruned. This prevents an uncontrolled growth in the number of Gaussians.

\begin{figure}[ht]
  \centering
  \includegraphics[width=\columnwidth]{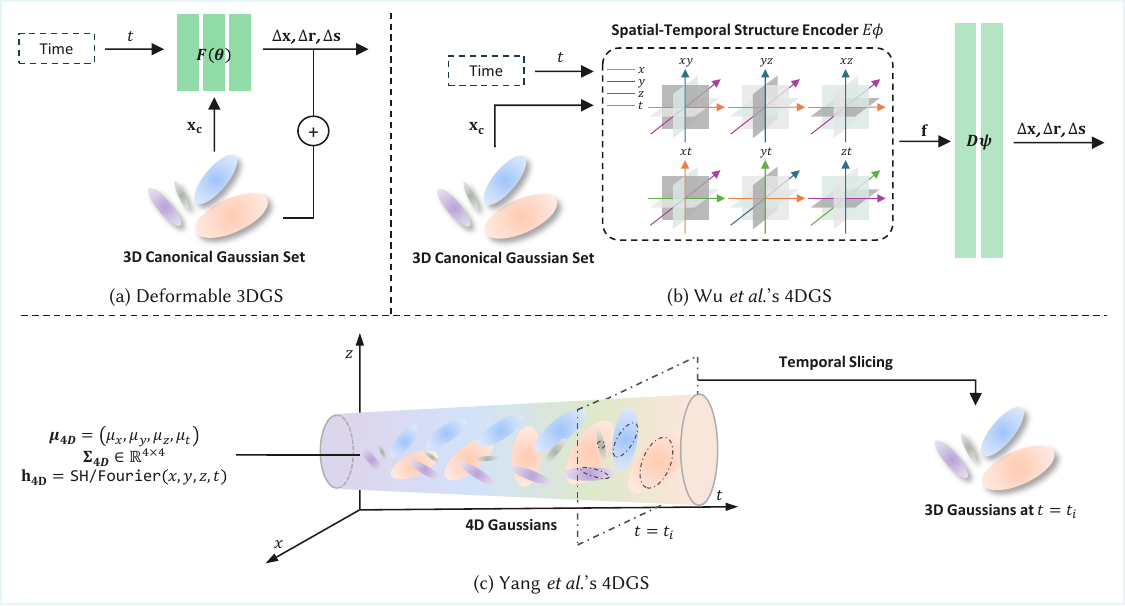}
  
  \caption{\textbf{Overview of representative approaches for dynamic 3D/4D Gaussian Splatting.} (a) \textbf{Deformable 3DGS}~\cite{b8} initializes a canonical set of 3D Gaussians and models temporal motion using a time-conditioned MLP that predicts offsets in position, rotation, and scale. (b) \textbf{Wu \textit{et al}.\,’s 4DGS}~\cite{b2} adopts a structured encoder–decoder design, where a spatial-temporal structure encoder (HexPlane) organizes the 4D domain $(x,y,z,t)$ into multi-plane features, which are then decoded into Gaussian attribute offsets. (c) \textbf{Yang \textit{et al}.\,’s 4DGS}~\cite{b43} directly extends 3D Gaussians into the spatio-temporal domain by defining each Gaussian as a 4D primitive with a 4D mean, covariance, and appearance coefficients. During rendering, 4D Gaussians are converted into 3D Gaussians at target timestamps through temporal slicing.}
  \label{figure2}
\end{figure}

\subsection{Dynamic 3D/4D Gaussian Splatting} \label{Sec. 2.2}
In recent years, research interest has shifted beyond static scene reconstruction using 3D Gaussian Splatting toward the reconstruction of dynamic scenes that incorporate a temporal dimension. As this field has gained momentum, various approaches have been proposed to extend 3D Gaussians for modeling dynamic scenes. In this section, we classify and describe how different dynamic 3DGS and 4DGS studies represent dynamic scenes. Representative approaches for dynamic 3D/4D Gaussian Splatting are illustrated in Fig.~\ref{figure2}. It should be noted that many works use the terms \textbf{dynamic 3DGS}~\cite{b10} and \textbf{4DGS}~\cite{b2, b43} interchangeably, and there is ongoing debate regarding which term to adopt. In this survey, we do not distinguish between the two terms.

Some models adopt deformation modeling with a canonical 3D Gaussian set. 
\textbf{Deformable 3DGS}~\cite{b8} initializes all 3D Gaussians in a canonical state and models their dynamic motion with a deformation network. 
The deformation network is a time-conditioned MLP that produces offsets for position, rotation, and scale:
\begin{equation}
(\Delta \mathbf{x}, \Delta \mathbf{r}, \Delta \mathbf{s}) = F_\theta(\mathbf{x}_c, t),
\end{equation}
where $\mathbf{x}_c$ is the canonical position and $t$ is the timestamp.

This design provides flexibility in modeling non-rigid motion and allows monocular dynamic scene reconstruction. 
However, since it relies on a dense MLP, the approach can suffer from temporal jitter, especially when supervision is limited. 
To mitigate this issue, Deformable 3DGS introduces Annealing Smooth Training (AST), which gradually regularizes the deformation field during optimization and effectively stabilizes the temporal dynamics.

\textbf{Wu \textit{et al}.\,'s 4DGS}~\cite{b2} extends the canonical Gaussian formulation with a structured encoder-decoder design. 
Instead of directly predicting offsets with a simple MLP, a spatial-temporal structure encoder organizes the 4D domain $(x,y,z,t)$ into multi-plane features. 
These features are interpolated to produce a latent code:
\begin{equation}
\mathbf{f} = E_{\phi}(\mathbf{x}_c, t),
\end{equation}
where $E_{\phi}$ denotes the HexPlane~\cite{b61} encoder. The latent feature $\mathbf{f}$ is then decoded into Gaussian attribute offsets:
\begin{equation}
(\Delta \mathbf{x}, \Delta \mathbf{r}, \Delta \mathbf{s}) = D_{\psi}(\mathbf{f}),
\end{equation}
where $D_{\psi}$ is a lightweight MLP decoder network. This grid-like representation improves expressiveness and motion fidelity compared to MLP-only approaches, 
but it requires more memory and results in slower rendering speed due to the cost of querying large spatio-temporal feature grids.

\textbf{Yang \textit{et al}.\,'s 4DGS}~\cite{b43} directly extends 3DGS into the spatio-temporal domain by defining each Gaussian as a four-dimensional primitive in $(x, y, z, t)$. 
In this formulation, time $t$ is not treated as a conditional input but rather as an independent coordinate dimension. 
Each Gaussian is parameterized with a 4D mean, a 4D covariance, and 4D appearance coefficients:

\begin{equation}
\boldsymbol{\mu}_{4D} = (\mu_x, \mu_y, \mu_z, \mu_t), \quad 
\boldsymbol{\Sigma}_{4D} \in \mathbb{R}^{4 \times 4}, \quad
\mathbf{h}_{4D} = \texttt{SH/Fourier}(x, y, z, t).
\end{equation}

Here, $\boldsymbol{\mu}_{4D}$ denotes the 4D position, 
$\boldsymbol{\Sigma}_{4D}$ is the 4D covariance matrix (encoding both scale and rotation), 
and $\mathbf{h}_{4D}$ represents spherical harmonics or Fourier coefficients defined over the 4D coordinates.
During rendering, 4D Gaussians must be converted into valid 3D Gaussians corresponding to the target timestamp $t_i$. 
This is achieved through temporal slicing, where each 4D Gaussian is projected onto the hyperplane $t = t_i$.
Unlike deformation network approaches, this method represents Gaussians as intrinsic 4D primitives that evolve continuously in time, 
allowing it to capture complex and periodic dynamics. 
For photorealistic rendering, Fourier basis functions are incorporated into the appearance coefficients to model high-frequency temporal variations.

%% file: sections/Sec.3.static.tex
\section{STATIC}
\label{static}
\textbf{Static 3DGS} requires a significant number of 3D Gaussians $\boldsymbol{\mathcal{G}}$ to maintain high fidelity.
This leads to large memory and storage requirements (often over 1GB for large real-world scenes~\cite{b6}).
This makes deployment difficult in resource-constrained environments like portable devices or head-mounted displays~\cite{b40}.
Therefore, reducing storage space and memory usage is essential to improve the scalability of 3DGS applications~\cite{b24}.
To this end, efficient Static 3DGS approaches can be categorized into two main directions:
\begin{itemize}
    \item \textbf{Parameter Compression}: These approaches compress the 3D Gaussians without modifying the original 3DGS~\cite{b1} model architecture.
    \item \textbf{Restructuring Compression}: These approaches fundamentally modify the original 3DGS~\cite{b1} model architecture to obtain an efficient scene representation.
\end{itemize}

\subsection{Parameter Compression} \label{Sec. 3.1}

Parameter Compression aims to reduce storage space and memory usage without modifying the 3DGS~\cite{b1} model architecture.
It can be applied to trained 3DGS models, making it flexible in various scenarios.
This survey classifies Parameter Compression methods into five main strategies:
\begin{itemize}
    \item \textbf{Pruning}: This strategy removes redundant or low-contribution 3D Gaussians based on various criteria.
    \item \textbf{Attribute Pruning}: This strategy simplifies Gaussian attributes by compressing specific components.
    \item \textbf{Quantization}: This strategy discretizes Gaussian attributes by reducing their bit precision.
    \item \textbf{Entropy Coding}: This strategy utilizes statistical redundancy in quantized attributes to minimize storage.
    \item \textbf{Structured Compression}: This strategy organizes 3D Gaussians by spatial relationships to improve efficiency.
\end{itemize}

\begin{figure}[ht]
  \centering
      \includegraphics[width=\columnwidth]{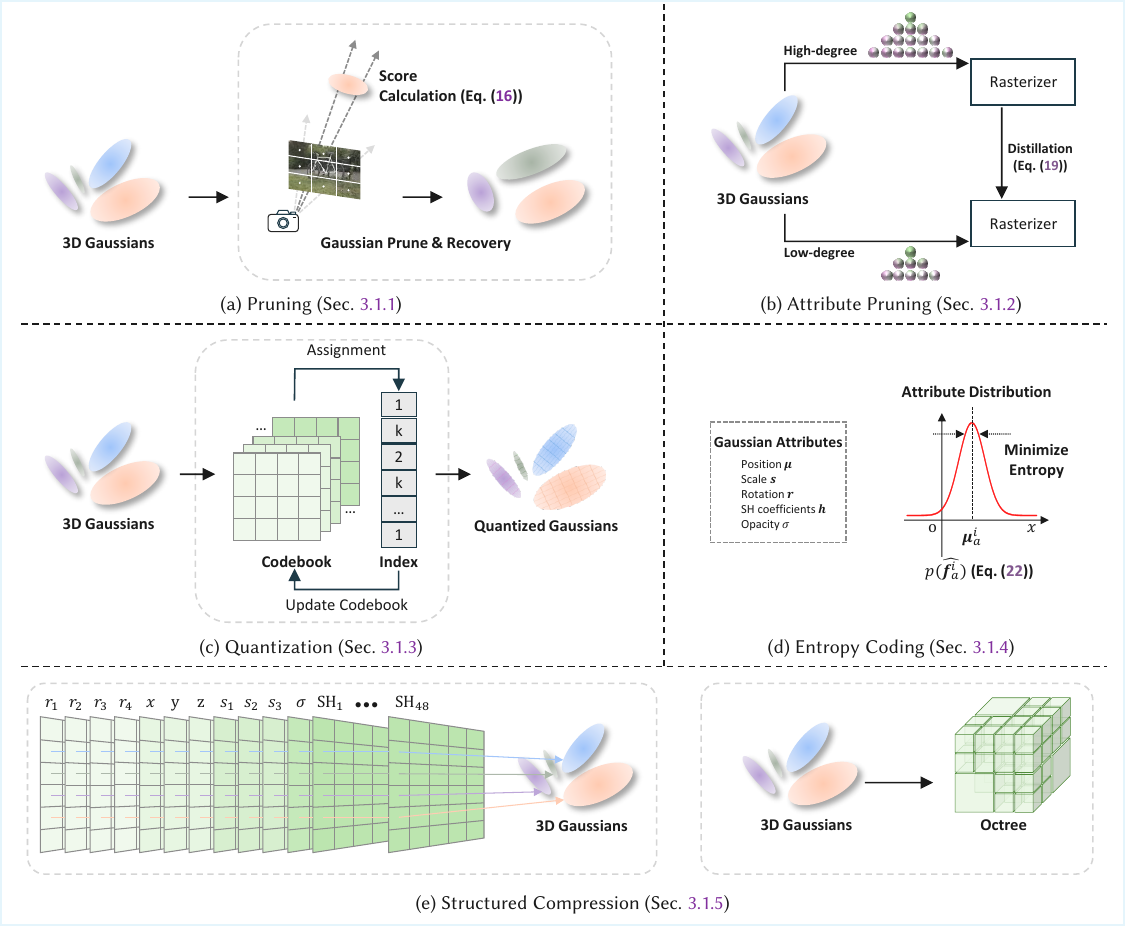}
  
  \caption{\textbf{Overview of Parameter Compression strategies for Static 3DGS}. These approaches reduce redundancy in the 3DGS representations without modifying the 3DGS~\cite{b1} model architecture. (a) \textbf{Pruning} (Sec.~\ref{subsubsec:static pruning}) removes redundant 3D Gaussians. (b) \textbf{Attribute Pruning} (Sec.~\ref{subsubsec:static attribute pruning}) compresses specific Gaussian attributes. (c) \textbf{Quantization} (Sec.~\ref{subsubsec:static quantization}) reduces Gaussian attribute bit precision. (d) \textbf{Entropy Coding} (Sec.~\ref{subsubsec:static entropy coding}) compresses quantized Gaussian attributes by exploiting statistical redundancy. (e) \textbf{Structured Compression} (Sec.~\ref{subsubsec:static structured compression}) organizes 3D Gaussians by spatial relationships to improve compression efficiency.}
  \label{figure3}
\end{figure}

\subsubsection{Pruning}
\label{subsubsec:static pruning}
\textbf{Pruning} is the most fundamental approach to reducing redundant 3D Gaussians.
To minimize the redundant 3D Gaussians, the original 3DGS~\cite{b1} periodically resets the opacity of 3D Gaussians to a low value at specific intervals and prunes Gaussians that remain below an opacity threshold after a certain number of iterations (Sec.~\ref{adaptive-density-control}).
However, according to Lee \textit{et al}.~\cite{b6}, relying solely on this opacity-based control method still results in a large number of redundant 3D Gaussians.
To this end, the pruning methods use various criteria to eliminate these redundancies.

~\hypertarget{learnable mask-based}{\textbf{Learnable mask-based Pruning}} uses learnable parameters to automatically select which 3D Gaussians to remove.
Lee \textit{et al}.~\cite{b6} use this approach to enable end-to-end learning of pruning decisions.
It considers both scale $\boldsymbol{s}$ and opacity $\sigma$ together, avoiding the fixed thresholds used in the original 3DGS~\cite{b1}.
The binary mask $\mathcal{M}_n \in \{0,1\}$ for the $n$-th 3D Gaussian is computed as follows through the learnable mask parameter $m_n \in [0,1]$:
\begin{equation}\label{eqn:(13)}
    \mathcal{M}_n = \operatorname{\texttt{sg}}(\operatorname{\mathds{1}}[\operatorname{\texttt{Sig}}(m_n) > \epsilon] - \operatorname{\texttt{Sig}}(m_n)) + \operatorname{\texttt{Sig}}(m_n)\ ,
\end{equation}
where $\operatorname{\texttt{sg}}(\cdot)$ denotes the Stop Gradient operator, $\mathds{1}[\cdot]$ denotes the indicator function, and $\operatorname{\texttt{Sig}(\cdot)}$ denotes the Sigmoid activation function.
The scale $\hat{\boldsymbol{s}}$ and opacity $\hat{\sigma}$ of the $n$-th masked 3D Gaussian are calculated as follows:
\begin{equation}\label{eqn:(14)}
    \hat{\boldsymbol{s}}_n = \mathcal{M}_n \boldsymbol{s}_n\ ,\ \hat{\sigma}_n = \mathcal{M}_n \sigma_n\ .
\end{equation}
Additionally, Lee \textit{et al}. regularize the model to minimize the redundant 3D Gaussians $\mathcal{\boldsymbol{G}}$ through masking loss $L_m$:
\begin{equation}\label{eqn:(15)}
    L_m = \frac{1}{\mathcal{N}} \sum_{n=1}^{\mathcal{N}} \operatorname{\texttt{Sig}}(m_n)\ .
\end{equation}

\textbf{Significance Score-based Pruning} methods introduce scoring functions to quantify the importance of each 3D Gaussian for scene reconstruction.
LightGaussian~\cite{b4} proposes a Global Significance (GS) score that comprehensively evaluates the contribution of each 3D Gaussian~\cite{b13}, inspired by the rendering equation Eq.~\eqref{eqn:(7)}.
As shown in Fig.~\ref{figure3}, the GS score for the $j$-th 3D Gaussian is calculated as follows:
\begin{equation}\label{eqn:(16)}
    GS_j = \sum_{i=1}^{|I| \times P_i} \mathds{1}[\boldsymbol{G}_j \land \boldsymbol{r}_i \neq \emptyset] \cdot \sigma_j \cdot \prod_{k=1}^{j-1} \left(1 - \sigma_k \right) \cdot \gamma(\boldsymbol{\Sigma}_j)\ ,
\end{equation}
where $|I|$ denotes the total number of images in the training set, and $P_i$ denotes the number of pixels in the $i$-th image.
The $\mathds{1}[\boldsymbol{G}_j \land \boldsymbol{r}_i \neq \emptyset]$ term represents whether the ray from the $i$-th pixel $\boldsymbol{r}_i$ intersects the $j$-th 3D Gaussian $\boldsymbol{G}_j$.
The $\gamma(\boldsymbol{\Sigma}_j)$ term represents the normalized volume of the $j$-th 3D Gaussian.
EAGLES~\cite{b13} introduces a new criterion to identify inefficient 3D Gaussians.
It calculates the influence $W_{i,p}$ of the $i$-th 3D Gaussian at a specific pixel $p$, and the total influence $W_i$ as follows:
\begin{equation}\label{eqn:(17)}
    W_{i,p} = \alpha_i \mathcal{T}_i = \alpha_i \prod_{j=1}^{i-1} (1 - \alpha_j)\ , \quad W_i = \sum_p W_{i,p}\ .
\end{equation}
MesonGS~\cite{b15} calculates the significance score for each 3D Gaussian by multiplying two components: a view-dependent significance score $W_i$ from Eq.~\eqref{eqn:(17)} and a volume-based view-independent significance score~\cite{b55, b56}.
OMG~\cite{b64} extends Eq.~\eqref{eqn:(17)} by additionally considering the following Local Distinctiveness:
\begin{equation}
    \text{Local Distinctiveness} = \left( \frac{1}{K} \sum_{j \in \boldsymbol{\mathcal{N}}_i^K} \|\boldsymbol{T}_i - \boldsymbol{T}_j\|_1 \right)^\lambda \ ,
\end{equation}
where $\boldsymbol{\mathcal{N}}_i^K$ denotes the set of the $k$-Nearest Neighbors (KNN) of the $i$-th 3D Gaussian and $\frac{1}{K} \sum_{j \in \boldsymbol{\mathcal{N}}_i^K} \|\boldsymbol{T}_i - \boldsymbol{T}_j\|_1$ represents the average difference in the appearance features $\boldsymbol{T}$.

\hypertarget{Gradient-based}{\textbf{Gradient-based Pruning}} methods calculate the importance of 3D Gaussians based on their gradient magnitudes.
PUP 3DGS~\cite{b38} measures 3D Gaussian significance using a sensitivity score derived from the diagonal components of the Hessian.
Speedy-Splat~\cite{b9} reparameterizes the Hessian approximation of PUP 3DGS to reduce storage requirements.
Trimming the Fat~\cite{b39} and ELMGS~\cite{b40} propose Gradient Aware Pruning (GAP), which uses both opacity and gradient to prune inefficient 3D Gaussians.

\subsubsection{Attribute Pruning}
\label{subsubsec:static attribute pruning}
\textbf{Attribute pruning} simplifies Gaussian attributes by selectively removing or compressing components that have a minimal impact on visual fidelity.
LightGaussian~\cite{b4} uses knowledge distillation to reduce the dimension of SH coefficients $\boldsymbol{h}$ while maintaining visual fidelity.
As shown in Fig.~\ref{figure3}, the $\texttt{student}$ model is trained to match the output of the $\texttt{teacher}$ model under the same camera pose $[\boldsymbol{R}|\boldsymbol{t}]$.
\begin{equation}
    \mathcal{L}_{\texttt{distill}} = \frac{1}{P_i} \sum_{i=1}^{P_i} \left\| \boldsymbol{C}_{\texttt{teacher}}(\boldsymbol{r}_i; [\boldsymbol{R}|\boldsymbol{t}]) - \boldsymbol{C}_{\texttt{student}}(\boldsymbol{r}_i; [\boldsymbol{R}|\boldsymbol{t}]) \right\|^2\ .
\end{equation}
RDO-Gaussian~\cite{b53} applies the masking strategy from Eq.~\eqref{eqn:(13)} to the certain Gaussian attributes.

\subsubsection{Quantization}
\label{subsubsec:static quantization}
\textbf{Quantization} discretizes Gaussian attributes by reducing their bit precision while maintaining visual fidelity~\cite{b56}.
Niedermayr \textit{et al}.~\cite{b7} define sensitivity $S(p)$ to quantify the impact of scalar attribute $p$ on total energy $E_j$ across the training images:
\begin{equation}\label{eqn:(24)}
    S(p) = \frac{1}{\sum_{i=1}^{|I|} P_i} \sum_{j=1}^{|I|} \left| \frac{\partial E_j}{\partial p} \right|\ .
\end{equation}
Based on Eq.~\eqref{eqn:(24)}, they introduce sensitivity-aware $k$-means clustering to the covariance matrix $\boldsymbol{\Sigma}$ and SH coefficients $\boldsymbol{h}$.
Lee \textit{et al}.~\cite{b6} use Residual Vector Quantization (R-VQ) to quantize Gaussian attributes.
LightGaussian~\cite{b4, b24} applies Vector Quantization (VQ) to the SH coefficients $\boldsymbol{h}$ with low GS scores, as defined in Eq.~\eqref{eqn:(16)}.
CompGS~\cite{b24} compresses Gaussian attributes using VQ.
RDO-Gaussian~\cite{b53} modifies the codeword selection process in VQ to jointly minimize rate loss $r_{i,k}$ and distortion loss $d_{i,k}$ for codeword index $k$.
For example, the codeword selection for the scale parameter $\boldsymbol{s}_i$ of the $i$-th 3D Gaussian is defined as follows:
\begin{equation}\label{eqn:(21)}
    j = \arg \min_k \frac{r_{i,k}^{(s)}}{\lambda^{(s)}} + d_{i,k}^{(s)} = \arg \min_k -\frac{\log p_k}{\lambda^{(s)}} + \left(s_i - \texttt{\textbf{CB}}^{(s)}[k]\right)^2\ .
\end{equation}
SizeGS~\cite{b56} introduces a size estimator which establishes a relationship between hyperparameters and compressed file size.
This relationship guides a Hierarchical Mixed Precision Quantization (H-MPQ) for Gaussian attributes, which applies different quantization precision levels to each attribute.
FlexGaussian~\cite{b159} proposes a training-free compression method that combines mixed-precision quantization and attribute-discriminative pruning.
This approach compresses 3DGS models without retraining.

\subsubsection{Entropy Coding}
\label{subsubsec:static entropy coding}
\textbf{Entropy coding} utilizes statistical redundancy in quantized Gaussian attributes to minimize storage requirements.
This strategy adaptively compresses a scene representation by utilizing statistical patterns.
Subsequent methods are particularly effective for 3DGS compression as quantized Gaussian attributes often exhibit high statistical redundancy.
Niedermayr \textit{et al}.~\cite{b7} use the DEFLATE algorithm~\cite{b93} to compress codebook indices.
RDO-Gaussian~\cite{b53} compresses the indices obtained from Eq.~\eqref{eqn:(21)} using arithmetic coding.
HAC~\cite{b30} models quantized anchor attributes $\hat{\boldsymbol{f}}_a$ as a Gaussian distribution $\phi_{\boldsymbol{\mu}_a, \boldsymbol{\sigma}_a}$ based on empirical observations.
Specifically, as shown in Fig.~\ref{figure3}, the probability of the quantized anchor features for the $i$-th anchor $p(\hat{\boldsymbol{f}}_a^i)$ is calculated as follows:
\begin{equation}
    \begin{aligned}
        p(\hat{\boldsymbol{f}}_a^i) &= \int_{\hat{\boldsymbol{f}}_a^i - \frac{1}{2}q_i}^{\hat{\boldsymbol{f}}_a^i + \frac{1}{2}q_i} \phi_{\boldsymbol{\mu}_a^i, \boldsymbol{\sigma}_a^i} (x) dx \\
        &= \Phi_{\boldsymbol{\mu}_a^i, \boldsymbol{\sigma}_a^i} \left( \hat{\boldsymbol{f}}_a^i + \frac{1}{2}q_i \right) - \Phi_{\boldsymbol{\mu}_a^i, \boldsymbol{\sigma}_a^i} \left( \hat{\boldsymbol{f}}_a^i - \frac{1}{2}q_i \right)\ , \\
        \boldsymbol{\mu}_a^i, \boldsymbol{\sigma}_a^i &= \texttt{MLP}_c (\boldsymbol{f}_h^i)\ ,
    \end{aligned}
\end{equation}
where $q_i$ denotes the adaptive quantization step size for the attribute of the $i$-th anchor, $\boldsymbol{f}_h^i$ represents the interpolated hash features, and $\texttt{MLP}_c (\boldsymbol{f}_h^i)$ is a Multi-layer Perceptron (MLP) that utilizes the interpolated hash features $\boldsymbol{f}_h^i$ to predict the mean $\boldsymbol{\mu}_a^i$ and standard deviation $\boldsymbol{\sigma}_a^i$ of the distribution of the quantized anchor attributes.
FCGS~\cite{b125} compresses 3DGS using a single forward pass, avoiding the scene-specific optimization process.
This framework utilizes a Multi-path Entropy Module (MEM) to compress Gaussian attributes, trading off size and visual fidelity.
Additionally, it uses Inter- and Intra-Gaussian Context Models to effectively eliminate redundancy among the irregular Gaussian blobs.

\begin{figure}[t]
    \centering
    \includegraphics[width=\columnwidth]{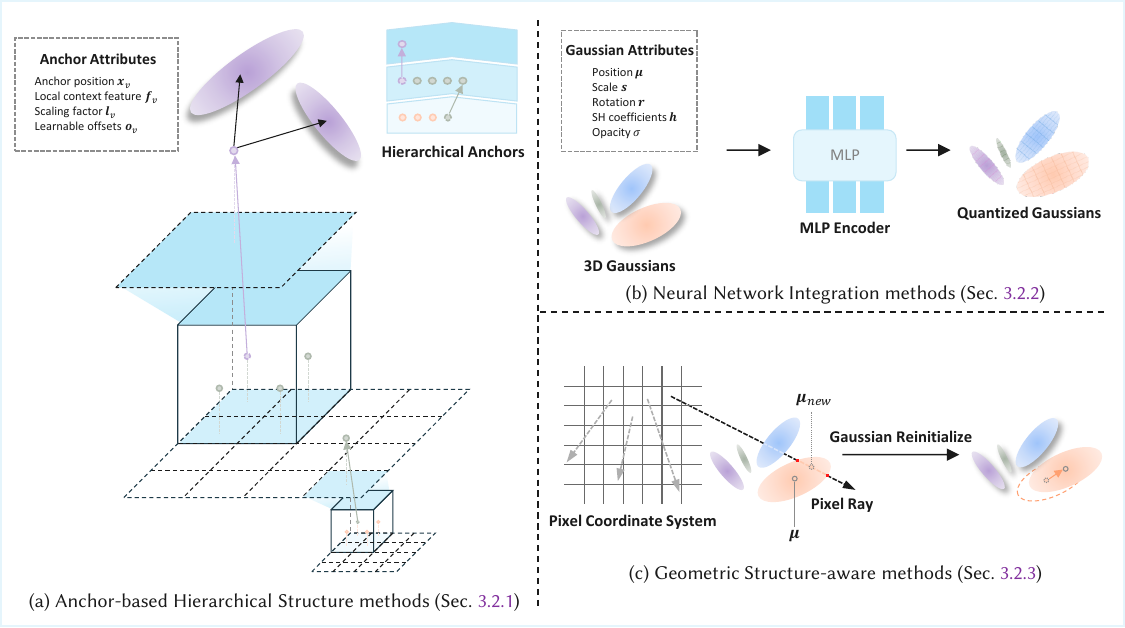}
     \caption{\textbf{Overview of Restructuring Compression strategies for Static 3DGS}. These approaches fundamentally modify the 3DGS model architecture to achieve an efficient scene representation. (a) Anchor-based Hierarchical Structure methods (Sec.~\ref{subsubsec:static anchor-based hierarchical structure methods}) introduce hierarchical representation to 3DGS~\cite{b1} using sparse anchors. (b) Neural Network Integration methods (Sec.~\ref{subsubsec:static neural network integration methods}) replace 3DGS representations with neural networks. (c) \textbf{Geometric Structure-aware} methods (Sec.~\ref{subsubsec:static geometric structure-aware methods}) exploit geometric properties.}
    \label{figure4}
\end{figure}

\subsubsection{Structured Compression}
\label{subsubsec:static structured compression}
\textbf{Structured Compression} addresses the compression inefficiency of the original 3DGS~\cite{b1}, where existing compression techniques~\cite{b71} achieve poor efficiency due to difficulty in identifying redundancy among unstructured 3D Gaussians.
These methods organize 3D Gaussians based on spatial relationships to improve compression efficiency.
Specifically, Morgenstern \textit{et al}.~\cite{b70} introduce the Parallel Linear Assignment Sorting (PLAS) algorithm to organize high-dimensional Gaussian attributes into a 2D grid, grouping spatially adjacent 3D Gaussians expected to have similar attributes.
Octree-GS~\cite{b37} utilizes an Octree~\cite{b75} structure to dynamically select Level-of-Detail (LoD) 3D Gaussians, enabling consistent real-time rendering for large-scale scenes.
Papantonakis \textit{et al.}~\cite{b67} use adaptive voxelization by computing the world-space volume of a single pixel and using 3D Gaussian overlap frequency within this volume as a spatial redundancy metric.
MesonGS~\cite{b15} utilizes an Octree structure to voxelize the 3D Gaussian positions $\boldsymbol{\mu}$ and uses RAHT (Region Adaptive Hierarchical Transform) to compress the Gaussian attributes in adjacent voxels. HGSC~\cite{b55} compresses 3D Gaussian positions $\boldsymbol{\mu}$ using an Octree structure and uses a multi-level attribute compression strategy. GoDe~\cite{b77} proposes a model-agnostic scalable compression framework that constructs a multi-level 3D Gaussian hierarchy, enabling dynamic adjustment of detail and compression rates without retraining. Wang \textit{et al.}~\cite{b80} introduce an adaptive voxelization algorithm utilizing transform coding tools developed for point cloud compression~\cite{b81}. GHAP~\cite{b129} compresses Gaussian attributes through block-wise Gaussian Mixture Reduction (GMR) based on KD-tree~\cite{b73} partitioning. HRGS~\cite{b158} introduces hierarchical block-level optimization, enabling high-quality, high-resolution 3D reconstruction even with limited GPU memory.

\subsection{Restructuring Compression} \label{Sec. 3.2}

Restructuring Compression aims to fundamentally modify the original 3DGS~\cite{b1} model architecture to obtain an efficient scene representation.
This survey classifies Restructuring Compression methods into three main strategies:
\begin{itemize}
    \item \textbf{Anchor-based Hierarchical Structure} methods: This strategy addresses the lack of a hierarchical representation in the original 3DGS~\cite{b1} by utilizing sparse anchor representations.
    \item \textbf{Neural Network Integration} methods: This strategy achieves an efficient scene representation by replacing the original 3DGS representations with neural networks.
    \item \textbf{Geometric Structure-aware} methods: This strategy achieves an efficient scene representation by utilizing scene geometric properties.
\end{itemize}

\subsubsection{Anchor-based Hierarchical Structure Methods}
\label{subsubsec:static anchor-based hierarchical structure methods}
\textbf{Anchor-based Hierarchical Structure} methods utilize the sparse anchor representations derived from the voxelization process as the structural foundation.
Scaffold-GS~\cite{b35} proposes a hierarchical scene representation based on anchors to address the lack of a hierarchical structure in the original 3DGS~\cite{b1}, which results in excessive redundant 3D Gaussians.
The framework uses the point clouds $\boldsymbol{P} \in \mathbb{R}^{N \times 3}$ to voxelize the scene as follows: $\boldsymbol{V} = \left\{ \operatorname{\texttt{Floor}} \left( \left| \frac{\boldsymbol{P}}{\epsilon} \right| \right) \right\} \cdot \epsilon$,
where $\epsilon$ is a value that represents the size of the voxel $\boldsymbol{V}$, and $\operatorname{\texttt{Floor}(\cdot)}$ is the floor function, which rounds down to the nearest smaller integer.
Each voxel $\boldsymbol{V}$ serves as an anchor $\boldsymbol{x}_v \in \boldsymbol{V}$, which is assigned the following attributes: local context features $\boldsymbol{f}_v \in \mathbb{R}^{32}$, scaling factors $\boldsymbol{l}_v \in \mathbb{R}^{3}$, and $k$ learnable offsets $\boldsymbol{O}_v \in \mathbb{R}^{k \times 3}$.
The framework produces $k$ neural Gaussians from each anchor $\boldsymbol{x}_v$ as follows:
\begin{equation}
    \{ \boldsymbol{\mu}_0, \ldots, \boldsymbol{\mu}_{k-1} \} = \boldsymbol{x}_v + \{ \boldsymbol{\mathcal{O}}_0, \ldots, \boldsymbol{\mathcal{O}}_{k-1} \} \cdot \boldsymbol{l}_v\ .
\end{equation}
Subsequently, the anchor attributes $\left\{ \boldsymbol{a}_0, \ldots, \boldsymbol{a}_{k-1} \right\}$ of each neural Gaussian are predicted using an MLP $F_{\boldsymbol{a}}$.
The MLP takes the view-dependent features $\hat{\boldsymbol{f}}_v$, the relative camera-anchor distance $\delta_{vc}$, and the view direction $\boldsymbol{d}_{vc}$ as input, as follows:
\begin{equation}
    \left\{ \boldsymbol{a}_0, \ldots, \boldsymbol{a}_{k-1} \right\} = F_{\boldsymbol{a}} \left( \hat{\boldsymbol{f}}_v, \delta_{vc}, \boldsymbol{d}_{vc} \right)\ .
\end{equation}
The view-dependent features $\hat{\boldsymbol{f}}_v$ is a weighted sum of multi-resolution local context features, where the weights depend on the relative distance $\delta_{vc}$ and the view direction $\boldsymbol{d}_{vc}$.
This approach enables the prediction of anchor attributes to be robust to changes in resolution and view direction.
HAC~\cite{b30} uses hash features derived from the interpolation of each anchor in a hash grid as contexts to estimate the probability of each quantized anchor attribute, enabling efficient entropy coding.
CompGS~\cite{b34} uses a hierarchical hybrid primitive structure that employs a small set of anchor primitives to predict the other primitives.
ContextGS~\cite{b12} divides the anchors into multiple hierarchical levels using a bottom-up voxelization strategy, as shown in Fig.~\ref{figure4}.
Subsequently, it introduces an autoregressive model that effectively predicts undecoded anchors by utilizing the already decoded anchors from the lower levels.
HEMGS~\cite{b84} achieves hybrid lossy-lossless compression of anchor attributes using a Hybrid Entropy Model (HEM).
The framework incorporates a variable-rate predictor for lossy compression and combines a hyperprior network with an autoregressive network for improved lossless entropy coding.
CAT-3DGS~\cite{b86} utilizes a Triplane-based hyperprior and a Spatial Autoregressive Model (SARM) to leverage the spatial inter-correlation among Gaussian primitives.
It introduces a Channel-wise Autoregressive Model (CARM) to utilize the channel-wise intra-correlation within individual primitives.
PCGS~\cite{b63} uses progressive masking to increase anchor quantity and progressive quantization with level-wise context modeling to refine anchor quality, enabling progressive bitstream generation.
TC-GS~\cite{b54} uses KNN to predict the distribution of Gaussian attributes based on the Tri-plane.
SHTC~\cite{b79} uses a Karhunen-Loève Transform (KLT)-based layer for data decorrelation and a sparsity-guided enhancement layer for residual compression.

\subsubsection{Neural Network Integration Methods}
\label{subsubsec:static neural network integration methods}
\textbf{Neural Network Integration} methods modify the original 3DGS~\cite{b1} model architecture by replacing the original 3DGS representations with neural networks.
This strategy utilizes the generative capacity of neural networks to compress Gaussian attributes.
EAGLES~\cite{b13} introduces an MLP decoder, denoted as $D: \mathbb{Z}^l \to \mathbb{R}^k$, which serves to decode the quantized latent vectors into actual attributes $\boldsymbol{\mathcal{A}}$.
As shown in Fig.~\ref{figure4}, the framework projects the uncompressed attributes $\boldsymbol{a} \in \boldsymbol{\mathcal{A}}$ into the latent space.
This projection applies the inverse decoder function $D^{-1}$ to obtain $\hat{\boldsymbol{q}} = D^{-1}(\boldsymbol{a})$.
The framework rounds $\hat{\boldsymbol{q}}$ to the nearest integer $\bar{\boldsymbol{q}}$ and reconstructs the attributes as $\boldsymbol{a} = D(\bar{\boldsymbol{q}})$.
NeuralGS~\cite{b157} performs clustering based on the similarity of Gaussian attributes and assigns a small MLP to each cluster to encode the Gaussian attributes.
Liu \textit{et al}.~\cite{b91} introduce the Mixture of Priors (MoP) and Coarse-to-Fine Quantization (C2FQ) strategies.
The MoP strategy uses multiple lightweight MLPs with a gating mechanism to generate diverse prior features, improving conditional entropy modeling accuracy.

\subsubsection{Geometric Structure-aware Methods}
\label{subsubsec:static geometric structure-aware methods}
\textbf{Geometric Structure-aware} methods modify the original 3DGS~\cite{b1} model architecture to utilize the geometric properties of the scenes.
This approach achieves a more efficient representation with high visual fidelity by incorporating geometric priors into a scene representation.
SAGS~\cite{b78} applies curvature-aware densification to COLMAP~\cite{b31} point clouds to populate under-represented areas.
It then introduces a structure-aware encoder that processes these densified points, utilizing Graph Neural Networks (GNNs) on a local-global graph representation to learn robust structural features.
Mini-Splatting~\cite{b11} reconstructs the spatial distribution of 3D Gaussians through densification with Blur Split and Depth Reinitialization.
Specifically, Depth Reinitialization determines the Gaussian depth as the mid-point of the two intersection points between the ray of each pixel and a 3D Gaussian, as shown in Fig.~\ref{figure4}.
These new points are reprojected into world space, contributing to the improvement of the inefficient spatial distribution.
3D-HGS~\cite{b122} addresses the limitations of the original 3DGS~\cite{b1} in modeling shape and color discontinuities by introducing a novel 3D Half-Gaussian (3D-HGS) kernel.
The kernel efficiently captures higher frequency information by using different opacity values for each half.

%% file: sections/Sec.3.dynamic.tex
\section{DYNAMIC}
\label{dynamic}
As 3DGS~\cite{b1} has shown remarkable success in static scene representation and real-time rendering, a natural yet challenging extension lies in modeling dynamic scenes. Real-world environments often contain non-rigid motions, occlusions, and temporally varying geometry and appearance. However, most early 3DGS methods are designed under the static scene assumption, which makes them unsuitable for applications such as video-based rendering, AR/VR, or autonomous driving. 

To bridge this gap, recent approaches~\cite{b10,b16} have extended the 3DGS framework to handle temporal dynamics. The primary challenge in this domain is achieving a compact and efficient representation that captures temporal variations while keeping computational and memory overhead low. Specifically, dynamic scenes demand additional storage to represent temporal variations across time~\cite{b17}, which scales linearly or exponentially depending on the granularity and duration of motion being modeled.

To address this, existing methods that address dynamic 3DGS typically fall into two broad categories. Following the same taxonomy introduced in the static setting, we analyze dynamic 3DGS methods through the lens of two complementary directions: \textbf{(1) Parameter Compression} and \textbf{(2) Restructuring Compression}. This dichotomy serves as a unified framework in our survey to assess how efficiently different approaches represent both static and dynamic scenes under a common goal of reducing redundancy:
\begin{itemize}
    \item \textbf{Parameter Compression}: These approaches focus on reducing redundancy in the spatio-temporal domain using pruning, quantization, and entropy-based methods, independent of the model architecture.
    \item \textbf{Restructuring Compression}: These approaches reduce representation size by reusing shared structures across frames through architectural designs such as anchor-based modeling or learned deformation fields.
\end{itemize}

\subsection{Parameter Compression}\label{Sec. 4.1}
Parameter compression methods aim to reduce redundancy in the Gaussian representation without modifying the core rendering architecture. These techniques typically operate at the level of individual Gaussians or their attributes, making them lightweight and easily integrated into existing 3DGS pipelines. We categorize recent parameter compression approaches as the following subtypes:
\begin{itemize}
    \item \textbf{Gaussian Pruning}: This strategy eliminates low-contributing Gaussians based on temporal activity.
    \item \textbf{Attribute Pruning}: This strategy removes low-impact Gaussian attributes.
    \item \textbf{Quantization}: This strategy discretizes Gaussian parameters.
    \item \textbf{Entropy-based}: This strategy exploits entropy in dynamic 3DGS.
\end{itemize}

\begin{figure}[!t]
    \centering
    \includegraphics[width=\columnwidth]{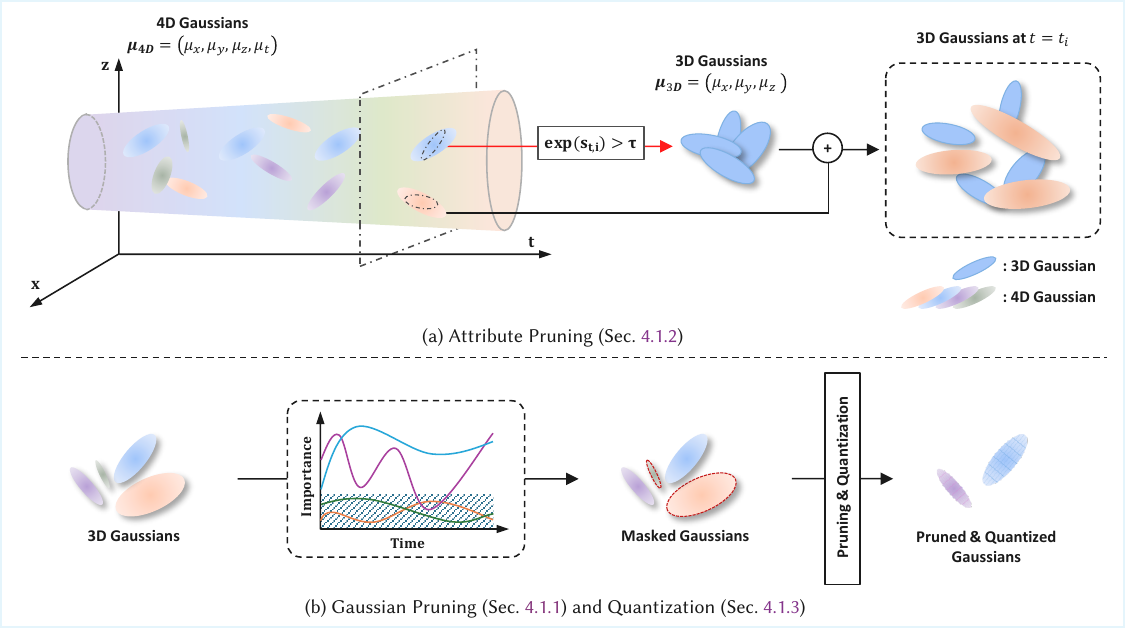}
     \caption{\textbf{Overview of Parameter Compression strategies for Dynamic 3DGS}. These approaches reduce redundancy in the Gaussian representation without modifying the rendering architecture. (a) \textbf{Attribute Pruning} (Sec.~\ref{subsubsec:attribute pruning}) removes temporally inactive components from 4D Gaussians, reducing them to 3D Gaussians that capture the invariant spatial attributes. The pruned 3D Gaussians are then combined with time-dependent 4D Gaussians at specific timestamps to reconstruct dynamic scenes while preserving spatial fidelity. (b) \textbf{Gaussian Pruning} (Sec.~\ref{subsubsec:gaussian pruning}) discards less-contributing Gaussians based on temporal importance, followed by \textbf{Quantization} (Sec.~\ref{subsubsec:quantization}), which discretizes Gaussian parameters to achieve compactness while preserving rendering quality.}
    \label{figure5}
\end{figure}

These methods serve as effective post-processing or training-time tools to trim over-parameterized models while maintaining high rendering quality, as summarized in Fig.~\ref{figure5}.

\subsubsection{Gaussian Pruning}
\label{subsubsec:gaussian pruning}
\textbf{Gaussian pruning} strategies have been widely explored in the field of static 3DGS to remove Gaussians with minimal visual impact~\cite{b4,b5,b6,b7}. These approaches typically assess importance based on spatial metrics or image-space contributions, which refers to the projected influence of a 3D Gaussian on the 2D image plane from a given camera viewpoint. However, extending such methods to dynamic scenes introduces additional challenges~\cite{b3,b23}. For instance, the pruning method, Lee \textit{et al}.~\cite{b6} described in static section~\ref{static} utilizes Eq.~\eqref{eqn:(13)} and~\eqref{eqn:(14)} to prune unnecessary Gaussians. Subsequently, a simple loss function~\eqref{eqn:(15)} is employed to encourage the mask parameter \(m_n\) to be as small as possible, thereby enabling effective pruning. In dynamic scenes, as Gaussians move over time, the importance of individual Gaussians can also change. This method does not account for temporal variations, which can lead to a Gaussian deemed important in one frame suddenly becoming unnecessary in the next, causing rendering quality degradation or unstable pruning. To address this, TC3DGS~\cite{b3} introduces a time attribute \(t\) and defines an additional mask consistency loss function:
\begin{equation}
    \mathcal{L}_{mc} = \sum_{n=1}^{N} |m_{n,t} - \operatorname{\texttt{sg}}(m_{n,t-1})|.
\end{equation}
This $\mathcal{L}_{mc}$ forces $m_{n,t}$ to remain similar to $m_{n,t-1}$, thereby ensuring that the importance of the Gaussians is learned in a temporally stable manner. This loss is then added to the basic loss~\eqref{eqn:(15)} to implement the final loss function. This allows the model to better grasp the global importance of Gaussians and effectively prune unnecessary ones, maximizing storage space and rendering efficiency.

Building upon the foundation described in~\hyperlink{Gradient-based}{static section}, Speedy-Splat~\cite{b9} computes a per-Gaussian sensitivity score by aggregating gradients across all static camera poses~$\phi$. SpeeDe3DGS~\cite{b8}, designed for dynamic scenes, extends this idea with the Temporal Sensitivity Pruning Score. Since Gaussians deform and change their contributions over time~$t$, it computes time-dependent gradients for the rendered image~$I_{\mathcal{G}_t}(\phi)$ and aggregates them across both camera poses and timestamps, enabling more stable and effective pruning over the dynamic sequence. In this direction, 4DGS-1K~\cite{b163} further advances pruning by jointly evaluating temporal and spatial scores. In particular, its temporal score quantifies the lifespan of each Gaussian along the time axis. By modeling the temporal opacity function \(p_i(t)\) and examining its second derivative, the method captures how steadily a Gaussian persists over time or how abruptly it appears or disappears, enabling more informed temporal pruning.

Also, Ex4DGS~\cite{b17} introduces a pruning strategy that leverages point backtracking, which traces image-space errors back to the responsible Gaussians. After computing pixel-wise errors by comparing rendered and GT images, the method exploits the backward pass to estimate each Gaussian’s contribution to the overall error. These contributions are weighted by opacity and accumulated transmittance to reflect their actual impact on pixels, and then averaged across all training views to obtain a global error threshold $\mathcal{E}_{total}$. Pruning is performed at predefined steps, where Gaussians with errors exceeding $\mathcal{E}_{total}$ are removed.


Another line of pruning-based strategies focuses on handling newly appearing objects while maintaining memory efficiency. 3DGStream~\cite{b161} assigns new Gaussians to emerging objects and applies Adaptive 3DG Quantity Control. When the view-space positional gradient, which measures how sensitively pixel colors change with respect to a Gaussian’s 2D projection, exceeds a threshold, new Gaussians are created, whereas those with gradients below the threshold are discarded. This allows the model to flexibly represent new content while preventing uncontrolled growth. Complementarily, Instant4D~\cite{b162} reduces redundancy and mitigates self-occlusion by partitioning the world space into a regular voxel grid and retaining only the centroid within each occupied voxel, keeping the representation compact during dynamic updates.

\subsubsection{Attribute Pruning}
\label{subsubsec:attribute pruning}
\textbf{Attribute pruning} focuses on removing specific attributes or parameters of Gaussians that are redundant or have minimal impact on the scene representation. A notable example is Hybrid 3D-4DGS~\cite{b21}, which prunes time attributes \(t\) from the 4D Gaussians. The core idea of Hybrid 3D-4DGS is a hybrid representation that models a scene by decomposing it into static and dynamic regions. It begins with a complete 4D Gaussian representation and then identifies Gaussians that do not change over time. These static 4D Gaussians are converted into 3D Gaussians, which effectively removes their temporal dimension parameters. This significantly reduces the total number of parameters, leading to lower memory consumption and faster training. Unlike prior works that often identify static and dynamic content by analyzing the flow of Gaussians~\cite{b17,b22,b24}, Hybrid 3D-4DGS leverages a 4D coordinate system. It uses a time-axis scale parameter, exp($s_{t,i}$), for each Gaussian. If this temporal scale exceeds a predefined threshold $\tau$, the Gaussian is classified as static and its time-attributes are pruned.

\subsubsection{Quantization}
\label{subsubsec:quantization}
\textbf{Quantization} is a core parameter compression technique that converts continuous, high-precision values into a discrete, low-precision representation. Vector Quantization (VQ) is a specific and powerful form of quantization that exploits the inherent redundancy in 3DGS parameters. It groups similar parameters into a few clusters, stores the cluster centers in a codebook, and then replaces the original parameters with the index of their corresponding cluster. This process significantly reduces storage, as millions of Gaussians can be represented with just the codebook and a set of indices. VQ is a widely used parameter compression technique in various applications, including deep network compression~\cite{b25, b26} and generative models~\cite{b27, b28, b29}. This demonstrates its proven efficacy in handling large-scale, redundant data, which is precisely the challenge posed by 3DGS. While initially developed for static scenes~\cite{b4, b6, b7, b24}, VQ is crucial for dynamic 3DGS due to the massive storage requirements of temporal data. The core principle remains the same, but the challenge lies in applying it to dynamic parameters, such as temporal deformation fields.

Not all Gaussian parameters equally affect rendering quality, making uniform quantization inefficient. Sensitivity-based quantization allocates higher bit precision to more critical parameters and lower precision to less important ones. In dynamic 3DGS, some parameters are more sensitive to temporal changes. TC3DGS~\cite{b3} addresses this with a gradient-aware mixed-precision scheme that measures each parameter’s sensitivity via its gradient magnitude. Parameters with larger gradients receive higher precision, while less impactful ones are quantized with fewer bits, balancing compression and reconstruction accuracy.

\subsubsection{Entropy-based}
The concept of \textbf{entropy} is leveraged in dynamic 3DGS in two ways. The first is as a compression codec used in the final compression stage after model optimization. The second is as a regularization loss during the training process to induce model sparsity. These two approaches stem from fundamentally different philosophies, and it's crucial to understand the distinct advantages and disadvantages of each.

Entropy is first used as a compression codec in the final compression stage. A key challenge with 4D Gaussians is that each frame consumes storage comparable to a keyframe, causing high memory usage for long sequences. HiFi4G~\cite{b97} addresses this with residual compensation, quantization, and entropy encoding. It exploits the small differences between adjacent frames by computing residuals between each Gaussian and its keyframe attributes, which are often near zero. These residuals are quantized and encoded using a Ranged Arithmetic Numerical System (RANS)~\cite{b98}, which efficiently compresses skewed frequency distributions. The encoded integer stream can be decoded to reconstruct the original attributes, enabling HiFi4G to achieve real-time compression and decompression even for long sequences.

The use of entropy as a regularization loss can be seen in MEGA \cite{b96}, which introduces an entropy-constrained Gaussian deformation technique to enhance the utilization of each Gaussian and reduce the total number required. Conventional Yang \textit{et al}.'s 4DGS method~\cite{b43} often assumes that 4D Gaussians exhibit only linear motion over time with constant covariance. Additionally, temporal decay opacity ensures that each Gaussian is only visible during a specific time, with less than 10\% of Gaussians participating in the rendering at any given moment. To  overcome these limitations, MEGA improves the flexibility of Gaussian motion and geometric structure. It then introduces a spatial opacity-based entropy loss to encourage each Gaussian's spatial opacity $\sigma$ to be close to 0 or 1:
\begin{equation}
    \mathcal{L}_{opa} = \frac{1}{N} \sum_{j=1}^{N} (-\sigma_j \log(\sigma_j)).
\end{equation}
Periodically during the optimization process, Gaussians with near-zero opacity are aggressively pruned, as they are considered to not be contributing to the scene's representation. This reduces the model's memory footprint and increases computational efficiency.

\subsection{Restructuring Compression} \label{Sec. 4.2}
While parameter compression methods reduce redundancy at the level of individual Gaussians or their attributes, restructuring compression techniques aim to achieve a more efficient representation by modifying the underlying architecture itself. These methods are \textbf{architecture-dependent} in that they introduce additional structural priors or modules that enable more efficient modeling of dynamic scenes. Specifically, recent restructuring compression methods can be broadly categorized into three major approaches:
\begin{itemize}
    \item \textbf{Anchor-based Representation}: This strategy compresses dynamic content using representative keyframes.
    \item \textbf{Canonical Deformable Representation}: This strategy models dynamic scenes by establishing a canonical space and a deformation field.
    \item \textbf{LoD Representation}: This strategy organizes Gaussians into multi-resolution hierarchies.
\end{itemize}

\begin{figure}[htbp]
  \centering
  \includegraphics[width=\linewidth]{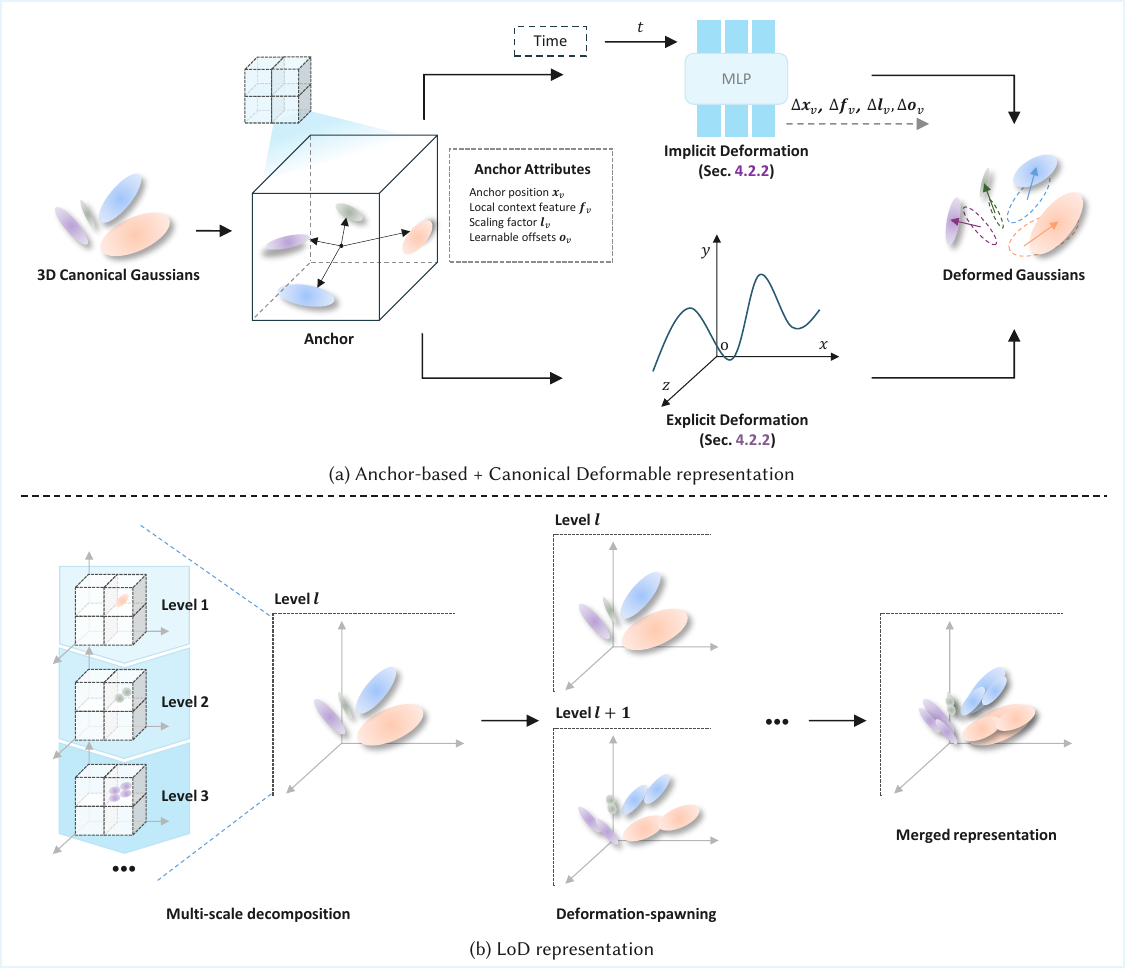}
  
  \caption{\textbf{Overview of Restructuring Compression strategies for Dynamic 3DGS.} These approaches achieve compact yet expressive representations by restructuring the underlying 3DGS architecture to better encode temporal and spatial variations. (a) \textbf{Anchor-based} and \textbf{Canonical Deformable Representation} (Sec. \ref{subsubsec:anchor-based representation}, Sec. \ref{subsubsec:canonical defromable representation}) combines anchor-based referencing with a canonical–deformable formulation. A small set of anchor Gaussians serve as spatial references, where each anchor encodes local context attributes and deformation parameters that map the canonical static space to dynamic frames over time. (b) \textbf{LoD Representation} (Sec. \ref{subsubsec:LoD representation}) organizes Gaussians into a multi-resolution hierarchy, progressively refining spatial detail through level-wise deformation and spawning}
  \label{figure6}
\end{figure}

These strategies reduce parameters, improve temporal consistency, and provide inductive biases that enhance generalization in dynamic environments, as summarized in Fig.~\ref{figure6}. In the following subsections, we review representative methods in each category. While we group them for clarity, recent works~\cite{b50, b58} often combine multiple techniques. Many state-of-the-art methods adopt hybrid designs that leverage the strengths of different paradigms, so these categories should be seen as conceptual guides rather than strict boundaries.

\subsubsection{Anchor-based Representation}
\label{subsubsec:anchor-based representation} Recent studies~\cite{b30, b34, b35, b12} have successfully introduced more compact 3DGS frameworks using anchor-based representations. This approach assigns implicit features to sparse anchor points, which are then used to predict the attributes for a broader set of neighboring 3D Gaussians. Building on this natural progression, anchor-based frameworks for static scenes are being actively extended to dynamic environments to address the significant storage overhead caused by the temporal dimension.

Anchor-based approach represents a dynamic scene's complex deformations by using a small number of hierarchical control points or anchors to represent a large number of Gaussians. These control points increase in density in areas with complex motion, allowing the model to effectively capture even subtle non-rigid deformations~\cite{b51}. MoSca~\cite{b57} separates the 3D geometry and motion of a dynamic scene using a representation called 4D Motion Scaffolds. These scaffolds compactly encode the underlying deformations of the scene, while 3D Gaussians are anchored to them to represent the appearance. This enables the model to globally encode information from all video frames into a single, consistent representation.
EDGS~\cite{b52} models dynamic scenes efficiently through a sparse anchor-grid representation. This approach decomposes time-invariant and time-variant properties. In particular, it employs an unsupervised learning strategy to effectively filter out anchors corresponding to static regions, while querying time-variant attributes by feeding only anchors related to dynamic objects into an MLP. Another notable method is MoDec-GS~\cite{b50}, introducing Global-to-Local Motion Decomposition (GLMD) to capture both global and local movements. This approach extends static scaffold representations to dynamic video reconstruction by using Global Canonical Scaffolds and Local Canonical Scaffolds.
HAIF-GS~\cite{b49} employs sparse motion anchors as deformation units to reduce redundant computation and improve efficiency. Its Hierarchical Anchors Densification (HAD) adaptively refines anchor resolution in motion-complex regions for fine-grained deformations. An Anchor Filter predicts dynamic confidence scores to suppress redundant updates in static areas, while the Induced Flow-Guided Deformation (IFGD) module aggregates multi-frame features to induce scene flow in a self-supervised manner, regularizing anchor transformations.

\subsubsection{Canonical Deformable Representation}\label{subsubsec:canonical defromable representation} This approach represents a dynamic scene as a set of static 3D Gaussians defined in a canonical space and a deformation field that maps points from this canonical space to the time-dependent real-world by implicitly or explicitly. Efficiency is achieved by inducing dynamic changes through the deformation field, rather than storing all Gaussians for every timestamp~\cite{b10}.

\textbf{Implicit Deformation} learns time-dependent deformations from the canonical space using MLP or grid-based architectures. Instead of defining a rigid, explicit function, the deformation field is modeled as a neural network that infers the transformation for each Gaussian at a given time. This allows for the capture of complex, non-rigid movements with high fidelity. One of the early works in this field, Deformable 3DGS~\cite{b42} introduces an implicit deformation field modeled by an MLP network, which, at each timestamp, takes a 3D Gaussian's position and the corresponding time $t$ as input to predict offsets for its position $\boldsymbol{\mu}$, orientation $\mathbf{r}$, and scale $\mathbf{s}$. Wu \textit{et al}.\,'s 4DGS~\cite{b2} takes a different approach by maintaining a single set of canonical 3D Gaussians and transforming them at each timestamp via a Gaussian deformation field. This field, comprising a temporal-spatial structure encoder and a multi-head Gaussian deformation decoder, leverages a multi-resolution HexPlane~\cite{b61}, a representation for 4D volumes that uses six planes of learned features to efficiently compute features for spatio-temporal points, thereby efficiently modeling Gaussian motion and shape changes. Building upon these, DeformGS~\cite{b60} focuses on dynamic 3D tracking using multiple cameras, employing a Neural-voxel Encoding combined with an MLP to infer a Gaussian's position, rotation, and even a shadow scalar. This scalar is a value between 0 and 1 that represents the shadow intensity. This approach further incorporates regularization terms based on conservation of momentum and isometry to reduce trajectory errors. 

\textbf{Explicit Deformation} models dynamic scenes using mathematically predefined functions to represent motion trajectories and temporal changes. Instead of relying on a neural network to predict transformations, this approach directly parameterizes motion with a set of explicit variables, which can include polynomial coefficients or keyframe-based interpolators. This strategy often leads to more compact representations by avoiding the need for a deep neural network. For example, Li~\textit{et al}.~\cite{b16} model dynamic scenes using an explicit deformation approach that achieves compactness through low-order polynomial functions. The position~$\boldsymbol{\mu}$ and orientation~$\mathbf{r}$ of each Gaussian are represented by these polynomials, with motion trajectories inspired by~\cite{b87,b1} and rotations by~\cite{b88,b89}. Representing time-varying motion with a small set of polynomial coefficients greatly reduces parameters compared to per-frame modeling, making the parameter count proportional to the number of Gaussians rather than frames. While other techniques like splatted feature rendering aid compactness, the core efficiency of Li~\textit{et al}.’s method lies in its explicit polynomial representation of dynamic motion. Another approach that uses explicit deformation is Ex4DGS~\cite{b17}, which efficiently represents the motion of dynamic 3DGS by employing keyframe-based interpolation. Instead of storing motion information for every frames, Ex4DGS explicitly stores the position and rotation of Gaussians only at a small number of keyframes. The scene's motion at any given timestamp is then synthesized by interpolating between adjacent keyframes. For the Gaussian's position $\boldsymbol{\mu}$, it uses Cubic Hermite Spline (CHip)~\cite{b94}, which provides a smooth and continuous interpolation
\begin{equation}
    \mu(t) = \operatorname{\texttt{CHip}}(p_n, m_n, p_{n+1}, m_{n+1}; t') .
\end{equation}
Here, $p_n$ is the position at the $n$-th keyframe and $m_n$ is its tangent vector, which is calculated based on the position change between keyframes to avoid extra storage. For the Gaussian's rotation, it uses Spherical Linear Interpolation (Slerp)~\cite{b95} to ensure a consistent and bias-free interpolation of the quaternion $\boldsymbol{q}$,
\begin{equation}
    q(t) = \operatorname{\texttt{Slerp}}(r_n, r_{n+1}; t') ,
\end{equation}
where $r_n$ is the rotation at the $n$-th keyframe. By using these explicit interpolators and keyframes, Ex4DGS avoids the need to store information for every single timestamp, which significantly reduces the model size. This approach effectively balances expressiveness for complex motion with the practicality of minimal memory overhead.

\subsubsection{LoD Representation}\label{subsubsec:LoD representation} As we described in the static section, \textbf{LoD representation} contributes to compact and efficient 3DGS frameworks by dynamically adjusting the level of detail of objects based on the viewer's perspective, distance, and importance. This makes it valuable for various applications like gaming~\cite{b103,b104} and virtual reality~\cite{b105,b106}.  Inspired by scalable video encoding~\cite{b107,b108}, Scale-GS~\cite{b109} employs LoD techniques with deformation and spawning to build a compact and efficient dynamic 3DGS framework. It introduces a multi-scale Gaussian representation, where large Gaussians capture coarse structures and small ones refine high-frequency details, improving efficiency by avoiding unnecessary optimization.
Scale-GS hierarchically partitions the scale space through recursive binary splitting. From the initial frame, it estimates each Gaussian’s scale and defines the maximum, minimum, and mean scales $(s_{min}^{(l)}, s_{max}^{(l)}, s_{mean}^{(l)})$ for each level~$l$. If a finer level~$l{+}1$ is required, the scale range is divided as:
\begin{equation}
[s_{\min}^{(l)}, s_{\max}^{(l)}] \leftarrow [s_{mean}^{(l)}, s_{\max}^{(l)}],
\end{equation}
\begin{equation}
[s_{\min}^{(l+1)}, s_{\max}^{(l+1)}] \leftarrow [s_{min}^{(l)}, s_{mean}^{(l)}].
\end{equation}
This hierarchy enables coarse-to-fine optimization, where large-scale Gaussians first approximate the global structure using low-resolution views, and smaller ones are later activated to refine fine details, greatly reducing redundant computation and training time. A similar hierarchical principle appears in 4DGCPro~\cite{b164}, which decouples motion into two levels: a rigid transformation capturing large object movements, and a residual deformation that models non-rigid shape changes and fine-grained local dynamics. This decomposition mirrors a coarse-to-fine strategy, allowing the system to first account for dominant global motion and then refine complex deformations efficiently.


%% file: sections/Sec.4.dataset.tex
\section{DATASETS AND EVALUATION}
\label{DatasetsAndEvaluation}
Evaluating efficient 3DGS and its dynamic extensions requires efficiency-oriented metrics beyond conventional rendering quality measures. While earlier studies focused primarily on visual quality (PSNR, SSIM~\cite{b33}, LPIPS~\cite{b117}), efficient approaches target a quality-efficiency trade-off. We summarize commonly used datasets and metrics from the literature and analyze the balance between visual fidelity and efficiency.

\subsection{Datasets}
\subsubsection{Static Scene Datasets}

\begin{table}[t]
\centering
\small
\begin{tabular}{lccccc}
\hline
 & \textbf{TNT}~\cite{b126} & \textbf{Deep Blending}~\cite{b127} & \textbf{NeRF-Synthetic}~\cite{b32} & \textbf{BungeeNeRF}~\cite{b128} & \textbf{Mip-NeRF 360}~\cite{b102} \\ \hline
\textbf{Venue} & ToG'17 & ToG'18 & ECCV'20 & ECCV'22 & CVPR'22 \\
\textbf{Type} & Real & Real & Synthetic & Mixed & Real \\
\textbf{Modality} & Multi-view & Multi-view & Multi-view & Multi-view & Multi-view \\
\textbf{\#Views} & 100--400 & 12--418 & 300 & 220--463 & 100--330 \\
\textbf{\#Scenes} & 14 (\textcolor{red}{10}+\textcolor{blue}{4}) & 19 (\textcolor{red}{14}+\textcolor{blue}{5}) & 8 (\textcolor{red}{0}+\textcolor{blue}{8}) & 12 (\textcolor{red}{12}+\textcolor{blue}{0}) & 9 (\textcolor{red}{5}+\textcolor{blue}{4}) \\
\textbf{Resolution} & 1920 $\times$ 1080 & 1228--2592 $\times$ 816--1944 & 800 $\times$ 800 & N/A & 4946 $\times$ 3286 \\
\textbf{Environment} & Mixed & Mixed & Indoor & Outdoor & Mixed \\
\textbf{Link} & \href{https://www.tanksandtemples.org/download/}{\faExternalLink*} & \href{http://visual.cs.ucl.ac.uk/pubs/deepblending/}{\faExternalLink*} & \href{https://drive.google.com/drive/folders/1cK3UDIJqKAAm7zyrxRYVFJ0BRMgrwhh4}{\faExternalLink*} & \href{https://drive.google.com/drive/folders/1ybq-BuRH0EEpcp5OZT9xEMi-Px1pdx4D}{\faExternalLink*} & \href{https://jonbarron.info/mipnerf360/}{\faExternalLink*} \\ \hline
\end{tabular}%
\caption{
\textbf{Summary of representative static 3DGS datasets}. Type indicates whether scenes are real-world or synthetic. Modality specifies the camera configuration (monocular vs. multi-view). \#Views denotes the number of viewpoints per scene. \#Scenes represents the total number of scenes, where \textcolor{red}{red} indicates outdoor and \textcolor{blue}{blue} indicates indoor scenes. Resolution indicates image dimensions in pixels (width $\times$ height). Environment describes the scene setting (indoor vs. outdoor).
}
\end{table}

As static scene reconstruction and novel view synthesis have established themselves as fundamental research areas in computer vision and graphics, numerous high-quality datasets have been developed to support and evaluate various methodologies. In this section, we review five representative static scene datasets widely used for 3D reconstruction. Each dataset presents unique challenges for evaluating model performance.

\textbf{Tanks and Temples (TNT)~\cite{b126}} introduces a benchmark dataset for evaluating large-scale 3D reconstruction techniques. The dataset comprises real-world captures acquired using an industrial laser scanner at submillimeter accuracy. The dataset captures videos primarily using gimbal-stabilized cameras (DJI Zenmuse X5R and Sony a7S II) at $4$K resolution. TNT includes $14$ diverse scenes with $100$--$400$ views each, categorized into intermediate ($8$ outdoor) and advanced ($4$ indoor, $2$ outdoor) groups, with images extracted at $1920 \times 1080$ resolution.

\textbf{Deep Blending~\cite{b127}} introduces a comprehensive dataset of $2,630$ real photographs from $19$ scenes. The dataset comprises captures from Chaurasia \textit{et al}.~\cite{b131} ($7$ scenes), Hedman \textit{et al}.~\cite{b132} ($4$ scenes), the Eth3D benchmark ($1$ scene), and newly captured scenes ($7$ scenes). Each scene contains $12$--$418$ images from different viewpoints, spanning 5 indoor and 14 outdoor environments with resolutions ranging from $1228 \times 816$ to $2592 \times 1944$ pixels.

\textbf{NeRF-Synthetic~\cite{b32}} presents a synthetic dataset of eight path-traced objects designed to evaluate novel view synthesis on complex geometry and realistic non-Lambertian materials. The dataset employs the Blender Cycles renderer for high-fidelity physically-based rendering. Each scene provides $100$ training views and $200$ test views. The eight objects (Chair, Drums, Ficus, Hotdog, Lego, Materials, Microphone, and Ship) exhibit complex geometry and intricate appearance properties. All scenes feature object-centric indoor setups in controlled studio environments with neutral backgrounds, rendered at $800 \times 800$ pixel resolution.

\textbf{BungeeNeRF~\cite{b128}} presents multi-scale datasets for novel view synthesis in extreme multi-scale scenarios. The collection includes synthetic data from Google Earth Studio~\cite{b32} and Blender, and real-world UAV captures. The dataset contains twelve outdoor cities (New York, San Francisco, Sydney, Seattle, Chicago, Quebec, Amsterdam, Barcelona, Rome, Los Angeles, Bilbao, and Paris), plus additional landscape and Blender-synthetic environments. Each city scene provides approximately $220$--$463$ multi-viewpoint images along the camera trajectory.

\textbf{Mip-NeRF360~\cite{b102}} introduces a real-world dataset for evaluating novel view synthesis in unbounded $360$-degree scenes. The dataset captures $9$ scenes with $100$--$330$ images per scene. The dataset comprises images captured using a Sony NEX C-3 for $5$ outdoor scenes (bicycle, flowers, garden, stump, treehill) and a Fujifilm X100V for $4$ indoor scenes (room, counter, kitchen, bonsai), with an average image resolution of $4946 \times 3286$ pixels.

\subsubsection{Dynamic Scene Datasets}

\begin{table}[t]
\centering
\small
\begin{tabular}{lccccc}
\hline
 & \textbf{Technicolor}~\cite{b113} & \textbf{D-NeRF}~\cite{b111} & \textbf{HyperNeRF}~\cite{b110} & \textbf{N3DV}~\cite{b112} & \textbf{NeRF-DS}~\cite{b114} \\ \hline
\textbf{Venue} & CVPR'17 & CVPR'21 & SIGGRAPH Asia'21 & CVPR'22 & CVPR'23 \\
\textbf{Type} & Real & Synthetic & Real & Real & Real \\
\textbf{Modality} & Multi-view & Multi-view & Monocular & Multi-view & Multi-view \\
\textbf{\#Views} & 16 & 100--200 & 1--2 & 18--21 & 2 \\
\textbf{\#Scenes} & 11 (\textcolor{red}{0}+\textcolor{blue}{11}) & 8 (\textcolor{red}{2}+\textcolor{blue}{6}) & 7 (\textcolor{red}{0}+\textcolor{blue}{7}) & 6 (\textcolor{red}{0}+\textcolor{blue}{6}) & 8 (\textcolor{red}{0}+\textcolor{blue}{8}) \\
\textbf{\#Frames} & 150--300 & 50--200 & 450--900 & 300 & 500 \\
\textbf{Resolution} & 2048 $\times$ 1088 & 800 $\times$ 800 & 1980 $\times$ 1080 & 2704 $\times$ 2028 & 480 $\times$ 270 \\
\textbf{Environment} & Indoor & Mixed & Indoor & Indoor & Indoor \\
\textbf{Link} & \href{https://www.interdigital.com/data_sets/light-field-dataset}{\faExternalLink*} & \href{https://www.albertpumarola.com/research/D-NeRF/index.html}{\faExternalLink*} & \href{https://hypernerf.github.io/}{\faExternalLink*} & \href{https://neural-3d-video.github.io}{\faExternalLink*} & \href{https://jokeryan.github.io/projects/nerf-ds/}{\faExternalLink*} \\ \hline
\end{tabular}%
\caption{
\textbf{Summary of representative dynamic 3DGS datasets}. Type indicates whether the dataset contains real-world captured scenes or synthetically generated scenes (Real vs. Synthetic). Modality specifies the camera configuration used for data acquisition (Monocular vs. Multi-view). \#Views denotes the number of camera viewpoints available per scene in the dataset. \#Scenes represents the total number of distinct scenes included in the dataset, where \textcolor{red}{red numbers} indicate outdoor scenes and \textcolor{blue}{blue numbers} indicate indoor scenes. \#Frames represents the number of frames per scene.
Since each dataset contains scenes of varying lengths, the frame count is expressed as an approximate range. Resolution indicates the video resolution in pixels (width $\times$ height) used for each dataset. Environment describes the setting where the scenes are captured or created (Indoor vs. Outdoor).
}
\end{table}

As dynamic 3DGS has become a popular research area, a wide variety of datasets have been introduced to support the rapid pace of development. In this section, we will review five representative datasets that are commonly used across the dynamic 3DGS literature previously discussed. These datasets are frequently cited as benchmarks for evaluating the performance of new methods.

\textbf{Technicolor Dataset}~\cite{b113} is a multi-view light-field video collection of various dynamic scenes, including close-ups of human faces and animated objects. It is captured using a synchronized $4\times4$ camera grid at 30 frames per second with a high resolution of $2048\times1088$ pixels. The dataset's precise synchronization and calibration make it a strong benchmark for validating fine details and complex textures in dynamic 3DGS methods. However, a key consideration is that each sequence has unique shift and calibration tables, requiring a specific post-processing pipeline for proper use.

\textbf{D-NeRF}~\cite{b111} is a synthetic extension of a static NeRF~\cite{b32} benchmark, designed for dynamic scenes. It features eight scenes with large deformation and non-Lambertian materials, which are surface whose color and brightness depend on the viewing direction as well as the light's direction. This enables the modeling of complex and realistic light interactions, such as those on glossy or reflective surfaces. The dataset is created by rendering 100--200 multi-view still frames per scene at $800\times800$ resolution and arranging them into sequences over time. Its primary advantage for 3DGS is that it is a clean dataset, free from real-world camera or geometric errors, making it ideal for evaluating an algorithm's core dynamic deformation modeling capabilities. However, the lack of real-world noise and variations limits its use for validating generalization to real-world multi-view videos.

\textbf{HyperNeRF}~\cite{b110} is a collection of video sequences captured with a single moving camera, specifically designed to include scene with topology changes, such as fluids, contacts, and separation, which are lacking in prior public datasets. The videos, each lasting 30--60 seconds, are subsampled to 15 fps and camera poses for all frames are estimated using COLMAP. For training, every fourth frame is used, with intermediate frames reserved for validation. From a 3DGS perspective, this dataset is valuable for testing the robustness of methods that handle topology changes and non-rigid deformations. However, its performance is sensitive to failures or ambiguities in single-camera pose registration, meaning that COLMAP's alignment quality and the frame sampling strategy directly impact the final result.

\textbf{Neural 3D Video Dataset}~\cite{b112} is a collection of real-world multi-camera videos that includes everyday indoor scenes, cooking, people, and challenging dynamic and optical effects like fire and steam. It also captures complex effects such as reflections, transparency, self-shadows, volumetric effects, and even topology changes like pouring liquids. This dataset is captured using 21 synchronized GoPro Black Hero 7 cameras, shooting at $2028\times2704$ resolution at 30 fps. Camera parameters are precisely estimated using COLMAP. Typically 18 views are used for training, and 1 view for qualitative and quantitative evaluations. From a 3DGS perspective, this dataset's strength lies in its high-resolution, multi-view nature, and the inclusion of challenging dynamic and optical phenomena. This makes it highly suitable for evaluating a model's real-world performance, as well as for comparing model compression and efficiency for long sequences. A noted limitation is the potential for color inconsistencies between views due to differences in camera color correction, which may require post-processing.

\textbf{NeRF-DS}~\cite{b114} is a real-world collection focused on dynamic specular objects, such as moving mirrors, metal, and glossy surfaces. It consists of eight scenes from everyday environments with various types of motion and deformation. This dataset is created specifically to address the scarcity of such dynamic specular cases in existing dynamic NeRF datasets. The data is captured using two rigidly mounted forward-facing cameras shooting simultaneously, with each scene containing two videos of approximately 500 frames. One video is used for training, and the other is used for testing. This design avoids the unrealistic "teleporting camera" problem of alternating between cameras. Camera poses are registered using COLMAP, with the stability of the SfM process improved by pre-applying a moving object mask obtained with MiVOS~\cite{b119}. From a 3DGS perspective, a key advantage of this dataset is that it enables evaluation of real-world challenges such as dynamic specular highlights and reflections, foreground--background separation, and the role of masking moving foregrounds during pose registration. It has been quantitatively shown that omitting such masks leads to significant errors, underscoring pose sensitivity and reflection-induced breakdowns of multi-view consistency.

\textcolor{black}{
\subsection{Evaluation Metrics}
To evaluate efficient 3DGS and its dynamic extensions, we consider both visual fidelity metrics and efficiency-oriented indicators. The following metrics are commonly adopted in the literature:
\begin{itemize}
    \item \textbf{PSNR (Peak Signal-to-Noise Ratio)}: PSNR quantifies the pixel-level accuracy between a rendered image and the GT reference. Higher PSNR values indicate lower reconstruction error and better fidelity.
    \item \textbf{SSIM (Structural Similarity Index Measure)~\cite{b33}}: SSIM measures perceptual similarity by considering luminance, contrast, and structural information between two images. A higher SSIM score represents a closer resemblance to the GT from a perceptual perspective.
    \item \textbf{LPIPS (Learned Perceptual Image Patch Similarity)~\cite{b117}}: LPIPS evaluates perceptual similarity using deep neural network features that approximate human visual judgment. Lower LPIPS values correspond to higher perceptual quality.
    \item \textbf{Model Size}: Model size is reported either in megabytes (MB) or in terms of the total number of Gaussians. A smaller model size indicates more compact representations and better memory efficiency.
    \item \textbf{Compression Ratio / Reduction Percentage}: These metrics measure the degree of compactness achieved compared to the original model. A higher compression ratio (or reduction percentage) reflects more effective elimination of redundancy while ideally preserving rendering quality.
    \item \textbf{Training Time}: It represents the total time required to optimize the model from initialization to convergence. Faster training time highlights the practicality of a method, particularly for large-scale or dynamic scenes.
    \item \textbf{Inference FPS (Frames Per Second)}: Inference speed is measured in FPS, indicating how efficiently a model can render frames in real time. Higher FPS values are crucial for interactive applications such as AR/VR and robotics.
\end{itemize}
}

\subsection{Quantitative and Visual Comparison of Gaussian Splatting Methods} As shown in Fig.~\ref{figure7} and Fig.~\ref{figure8}, we visualize representative static and dynamic 3DGS methods using bubble charts that jointly consider reconstruction quality (PSNR ↑, LPIPS ↓), rendering speed (FPS ↑), and model size (bubble radius). For fair and comprehensive visualization, Mip-NeRF 360~\cite{b102} is adopted for static scenes and N3DV~\cite{b112} for dynamic ones, as these are the most widely used benchmarks in each category.

\subsubsection{Static 3DGS}
\begin{figure}[ht]
    \centering
    \includegraphics[width=0.7\columnwidth]{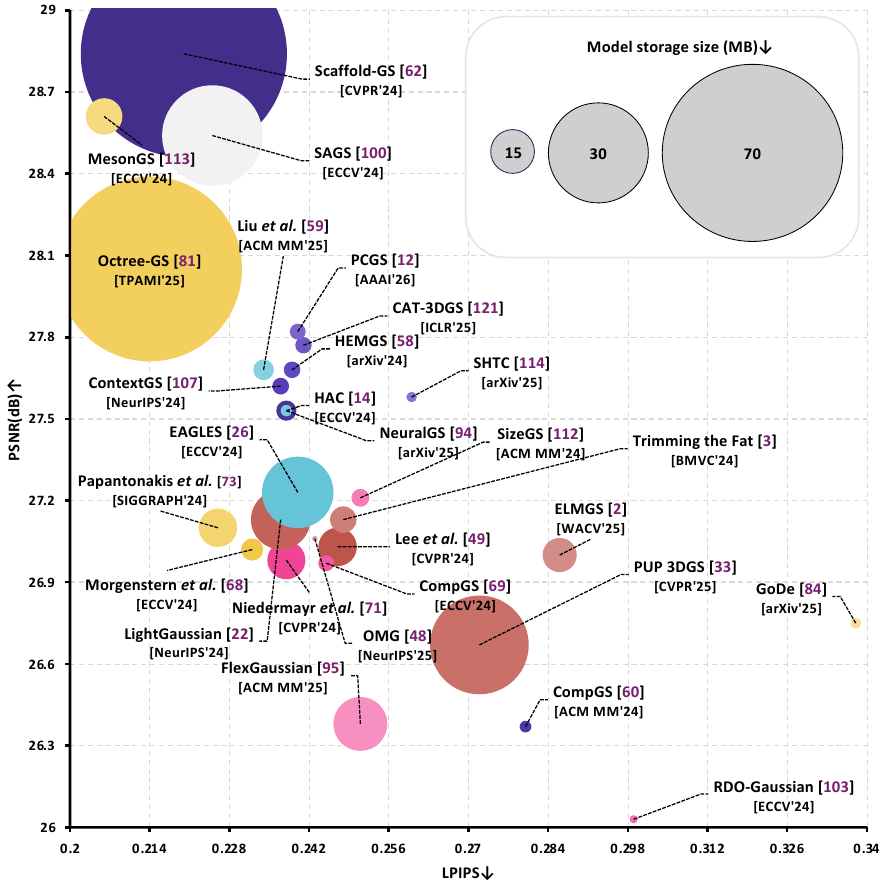}
     \caption{\textbf{Performance comparison visualization graph}. Bubble plot comparing static 3DGS methods on Mip-NeRF 360~\cite{b102}. The x-axis shows perceptual quality (LPIPS ↓), the y-axis shows reconstruction quality (PSNR in dB ↑), and the bubble radius is proportional to model storage size (MB ↓). For clarity, various compactness-oriented models are distinguished by color tones: Brick red for Pruning methods (Sec.~\ref{subsubsec:static pruning}), raspberry for Quantization methods (Sec.~\ref{subsubsec:static quantization}), vivid yellow for Structured Compression methods (Sec.~\ref{subsubsec:static structured compression}), deep violet for Anchor-based Hierarchical Structure methods (Sec.~\ref{subsubsec:static anchor-based hierarchical structure methods}), sky blue for Neural Network Integration methods (Sec.~\ref{subsubsec:static neural network integration methods}), and light gray for Geometric Structure-aware methods (Sec.~\ref{subsubsec:static geometric structure-aware methods}). Several compact designs achieve improved PSNR and lower LPIPS with reasonable storage, demonstrating the shift toward lightweight yet high-fidelity representations.}
    \label{figure7}
\end{figure}

As illustrated in Fig.~\ref{figure7}, the performance differences among static 3DGS compression methods are closely associated with the key technical features adopted by each approach. MesonGS~\cite{b15}, which demonstrates superior performance in the upper-left region of the chart, effectively exploits spatial adjacency through voxelization with an Octree structure~\cite{b75} and RAHT-based attribute compression. Similarly, Octree-GS~\cite{b37} achieves both high visual quality and efficiency in large-scale scenes due to its hierarchical structure that enables dynamic LoD selection. Scaffold-GS~\cite{b35} and SAGS~\cite{b78} provide the highest visual quality due to anchor-based hierarchical representation and structural feature learning via GNNs, respectively, although these complex structures require large model sizes. ELMGS~\cite{b40} and Trimming the Fat~\cite{b39} maintain competitive visual quality even with extremely small model sizes due to their adoption of gradient-based aggressive pruning strategies. ContextGS~\cite{b12} and HAC~\cite{b30} achieve an efficient quality-size tradeoff by exploiting spatial correlations through autoregressive models and hash-based entropy coding, respectively. HEMGS~\cite{b84} achieves high visual quality at low bitrates through hybrid entropy models that enable lossy-lossless compression. CAT-3DGS~\cite{b86} exploits inter-channel correlations via Triplane-based hyperpriors to achieve high visual quality. EAGLES~\cite{b13} and NeuralGS~\cite{b130} achieve smaller model sizes through MLP decoders and clustering-based encoding, respectively, by directly utilizing the generative capabilities of neural networks for compression. Liu \textit{et al}.~\cite{b91} maintain relatively high visual quality by improving the accuracy of conditional entropy modeling through a MoP strategy. CompGS~\cite{b24} and Niedermayr \textit{et al}.~\cite{b7} apply sensitivity-based $k$-means clustering, while Lee \textit{et al}.~\cite{b6} apply Residual Vector Quantization, both achieving simple but effective compression. LightGaussian~\cite{b4} achieves effective compression of SH coefficients through a combination of knowledge distillation and VQ, enabling substantial size reduction while maintaining visual fidelity. Papantonakis \textit{et al.}\cite{b67} and Morgenstern \textit{et al}.\cite{b70} effectively eliminate spatial redundancy through adaptive voxelization and the PLAS algorithm, respectively, resulting in compact representations. CompGS~\cite{b34} achieves efficient compression by exploiting predictive relationships between anchor and non-anchor primitives through a hierarchical hybrid primitive structure. PUP 3DGS~\cite{b38} attains smaller model sizes by selectively removing 3D Gaussians with low importance through Hessian-based sensitivity scores. OMG~\cite{b64} improves pruning effectiveness by performing pruning that considers local distinctiveness. SizeGS~\cite{b56} achieves predictable compression ratios by explicitly modeling the hyperparameter-size relationship via a Size Estimator to guide mixed-precision quantization. GoDe~\cite{b77} provides scalable quality control by enabling dynamic detail adjustment through progressive hierarchical structure construction via gradient-informed masking. PCGS~\cite{b63} enables progressive decoding by supporting progressive bitstream generation through progressive masking and level-wise context modeling. SHTC~\cite{b79} achieves efficient residual compression through the combination of KLT-based decorrelation and a sparsity-guided enhancement layer, maintaining high quality at reduced bitrates. FlexGaussian~\cite{b159} demonstrates competitive performance without retraining due to the combination of training-free mixed-precision quantization and attribute-wise pruning, offering practical deployment advantages. RDO-Gaussian~\cite{b53} maximizes compression efficiency by directly integrating rate-distortion optimization into the codeword selection process.

\subsubsection{Dynamic 3DGS} 
\begin{figure}[ht]
    \centering
    \includegraphics[width=0.7\columnwidth]{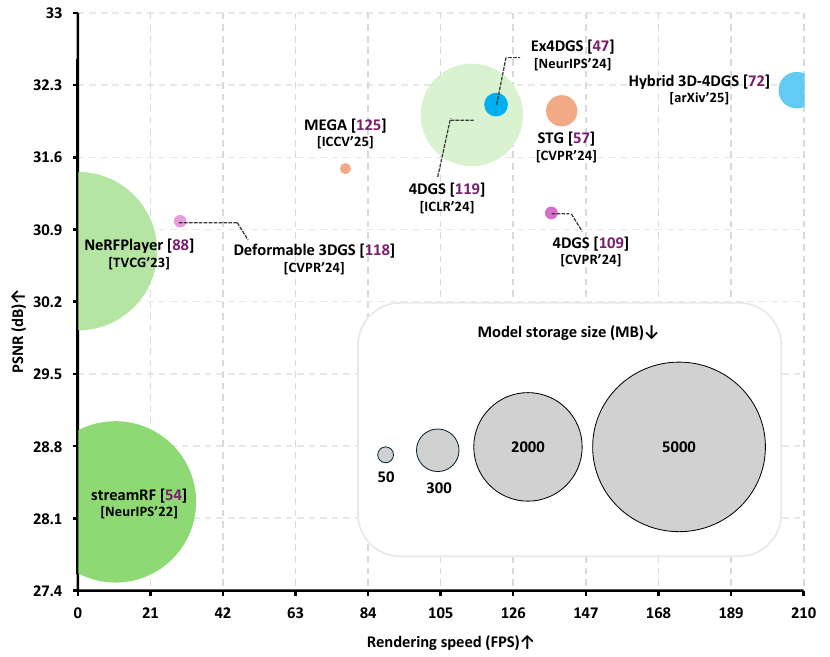}
     \caption{\textbf{Performance comparison visualization graph}. Bubble plot comparing dynamic 3DGS methods on N3DV~\cite{b112}. The x-axis shows rendering speed (FPS ↑), the y-axis shows reconstruction quality (PSNR in dB ↑), and the bubble radius is proportional to model storage size (MB ↓). For clarity, both compactness-oriented models (in blue tones) and non-compact baselines are shown. A clear reduction in bubble size can be observed among compact models, indicating a substantial decrease in storage cost. Moreover, several compact designs achieve improved PSNR while maintaining high rendering speed, revealing the direction of dynamic 3DGS research toward lightweight yet high-fidelity representations.}
    \label{figure8}
\end{figure}

As illustrated in Fig.~\ref{figure8}, a few NeRF-based methods~\cite{b136, b137} and non-compact 4DGS~\cite{b43} are also included for reference, where 4D Gaussian primitives achieve higher PSNR and FPS by modeling geometry explicitly. Deformable 3DGS~\cite{b42} provides one of the earliest compression strategies by avoiding the need to store all Gaussian attributes for every frame. Instead, it adopts a canonical space with a learned deformation field, yielding a substantial reduction in model size compared to per-frame representations. Building on this idea, 4DGS~\cite{b2} further improves efficiency by incorporating HexPlane-based decomposed neural voxel encoding, which results in notably higher rendering speed. Subsequent works explore selective dynamic modeling. Ex4DGS~\cite{b17} separates static and dynamic regions and leverages keyframe-based temporal interpolation with pruning to reduce memory footprint even further. Hybrid 3D–4DGS~\cite{b21} similarly performs static–dynamic decomposition, while retaining full 4D expressiveness for dynamic Gaussians to faithfully capture complex motions. Another line of research focuses on compressing spherical harmonics (SH), the dominant contributor to storage cost. STG~\cite{b16} and MEGA~\cite{b96} reduce SH complexity to obtain compact representations, with MEGA further benefiting from entropy-based regularization.


%% file: sections/Sec.6.limitations_and_future_directions.tex
\section{LIMITATIONS AND FUTURE DIRECTIONS} \label{Sec. 6}
Efficient static 3DGS and dynamic 3DGS have overcome the issue of slow training and rendering speed inherent in NeRF, enabling real-time rendering and high-quality scene reconstruction.
However, their practical deployment remains challenging due to massive memory footprint and computational overhead.
A typical high-resolution static scene often contains millions of 3D Gaussians. Furthermore, 4D scene reconstruction demands even greater memory consumption because each 3D Gaussian must encode temporal information across multiple frames.
To address these challenges, this survey categorizes and analyzes approaches for improving the efficiency of both static 3DGS and dynamic 3DGS.
All research presented in this survey is categorized into two approaches: Parameter Compression and Restructuring Compression.

\textbf{Parameter Compression} approach aims to reduce redundant 3D Gaussians or their attributes without modifying the original 3DGS~\cite{b1} model architecture.
This makes it flexible, as it can be applied to already trained models.

For static 3DGS, parameter compression strategies can be categorized as follows:
\begin{itemize}
    \item \textbf{Pruning} removes redundant or low-contribution 3D Gaussians.
    \item \textbf{Attribute Pruning} selectively removes specific components with minimal impact on rendering quality.
    \item \textbf{Quantization} reduces the bit precision of Gaussian attributes to minimize storage.
    \item \textbf{Structured Compression} organizes 3D Gaussians using spatial adjacency or hierarchical relationships to enhance compression efficiency.
\end{itemize}

For dynamic 3DGS, parameter compression strategies can be categorized as follows:
\begin{itemize}
    \item \textbf{Gaussian Pruning} eliminates low-contributing 3D Gaussians based on temporal activity or motion magnitude.
    \item \textbf{Attribute Pruning} eliminates temporal parameters, effectively converting some 4D Gaussians into 3D ones.
    \item \textbf{Quantization} applies sensitivity-based quantization methods to temporal parameters.
    \item \textbf{Entropy-based Compression} employs entropy-based codecs.
\end{itemize}

\textbf{Restructuring Compression} fundamentally modifies the original 3DGS~\cite{b1} model architecture to obtain an efficient scene representation. This category achieves compression through architectural redesign, such as hierarchical structures or alternative primitive representations.

For static 3DGS, restructuring compression strategies can be categorized as follows:
\begin{itemize}
    \item \textbf{Anchor-based Hierarchical Method} uses sparse anchors to address the lack of hierarchical structure in the original 3DGS~\cite{b1}, reducing redundancy.
    \item \textbf{Neural Network Integration} utilizes neural networks to compress Gaussian attributes by learning compact latent representations.
    \item \textbf{Geometric Structure-aware Method} utilizes geometric properties of the scene to improve efficiency.
\end{itemize}

For dynamic 3DGS, restructuring compression strategies can be categorized as follows:
\begin{itemize}
    \item \textbf{Anchor-based Representation} models complex deformations through sparse anchors.
    \item \textbf{Canonical Deformable Representation} maps a canonical static space to time-varying space using a deformation field.
    \item \textbf{LoD Representation} builds multi-resolution hierarchies to adjust detail dynamically.
\end{itemize}

Despite these advances in efficient static 3DGS and dynamic 3DGS, several critical challenges remain unresolved. This section discusses these limitations and proposes future research directions.

\subsection{Hardware Optimization and Real-time Deployment}
For practical deployment of static and dynamic 3DGS, optimization across diverse hardware platforms is essential. Although most current efficient 3DGS methods for both static and dynamic scenes are developed and evaluated in high-performance GPU environments, real-world applications such as AR/VR headsets, mobile devices, and edge devices operate under constrained memory and computational resources. Therefore, compression techniques and rendering pipelines that account for hardware constraints must be developed.

\subsection{Long-sequence Processing for Dynamic Scenes}
Most current dynamic 3DGS research targets relatively short video sequences. However, real-world applications require processing long sequences. In such long sequences, memory requirements grow exponentially, and maintaining temporal consistency becomes challenging. To this end, novel compression strategies that effectively exploit temporal redundancy are required. This can be addressed by extending keyframe-based interpolation methods to perform adaptive keyframe selection, or introducing hierarchical temporal encoding structures capable of modeling long-range temporal dependencies.

\subsection{Semantically-aware Compression}
Current static and dynamic 3DGS compression techniques primarily focus on pixel-level reconstruction accuracy, without sufficiently exploiting the semantic properties of scenes. Real-world scenes contain various semantic categories such as objects, backgrounds, and materials, which can be utilized to achieve more efficient compression. Specifically, scenes can be semantically segmented to allocate higher bit budgets to important foreground objects while allocating lower bit budgets to less important backgrounds. Since 3D Gaussians belonging to the same semantic category are likely to share similar characteristics, category-specific codebooks can achieve more effective compression. In dynamic scenes, explicitly separating dynamic objects from static backgrounds and applying semantically-aware compression strategies is also promising.

\subsection{Generalization}
Most current static and dynamic 3DGS methods employ per-scene optimization, requiring training from scratch for each individual scene. This results in substantial computational cost and training time for every new scene. In contrast, recent studies such as VGGT~\cite{b166} can reconstruct 3D scenes from only a few images without per-scene optimization. However, these foundation models are computationally expensive and memory-intensive, making them impractical for resource-constrained environments such as mobile devices. Future research should focus on developing lightweight compression models that preserve the generalization capability of foundation models while being compact enough for mobile deployment. In this context, single forward pass compression frameworks without per-scene optimization, such as FCGS~\cite{b125}, are also promising.

\subsection{User-controllable Quality-efficiency Trade-offs}
Most current efficient static and dynamic 3DGS research targets fixed quality-efficiency trade-offs. However, real-world applications require dynamic adjustment of this trade-offs based on user requirements or execution environments. Therefore, flexible compression frameworks that enable diverse rate-distortion trade-offs without retraining are required. Recent studies such as GoDe~\cite{b77} have begun exploring this direction, but more precise control over rate-distortion trade-offs remains challenging.

\subsection{Reliability and Robustness Enhancement}
For efficient static and dynamic 3DGS models to be deployed in safety-critical applications, reliability and robustness must be ensured. Specifically, adaptive compression strategies that identify safety-critical regions and allocate higher bit budgets to 3D Gaussians in those areas are required. Furthermore, quantifying compression-induced uncertainty and applying LoD to enhance details in high-uncertainty areas can improve reliability.

%% file: sections/Sec.7.conclusion.tex
\section{CONCLUSION} \label{conclusion}
Although 3DGS enables real-time rendering through its explicit representation, its memory overhead remains one of the major challenges. Recent studies have therefore focused on designing compact and efficient representations that reduce memory usage while preserving high-fidelity scene reconstruction for both static and dynamic scenes. This survey reviews recent works that address this challenge by proposing compact and efficient 3DGS frameworks. We first introduce the core concepts of Gaussian splatting in the Preliminary section (Sec.\ref{Sec. 2}). Building on this foundation, the Static and Dynamic sections (Sec.\ref{static}, Sec.\ref{dynamic}) systematically review recent methods that pursue compact and efficient representations under consistent criteria. We then summarize widely used datasets and evaluation metrics to support fair benchmarking across studies (Sec.\ref{DatasetsAndEvaluation}), and conclude by discussing current limitations and promising future directions in 3DGS research (Sec.~\ref{Sec. 6}).

%% file: arXiv_supplement.tex
\section{NOTATION TABLE}

All notations and symbols appearing in this survey are grouped into five categories according to their usage in Tab.~\ref{tab:notation}. The table provides a concise reference for readers unfamiliar with common 3DGS notation, including all frequently used variables and those mentioned at least once. Vector quantities are highlighted in boldface for clarity.

\input{sections/Sec.5.notation}

%% file: sections/Sec.5.notation.tex
\begin{longtable}{p{0.25\textwidth}p{0.7\textwidth}}
\label{tab:notation} \\
\toprule
\rowcolor{gray!25} \textbf{Notation} & \textbf{Explanation} \\ 
\midrule

\multicolumn{2}{l}{\textbf{1. Basic Elements of 3D Gaussians}} \\
\rowcolor{gray!25} \textbf{$\boldsymbol{\mu}$} & Center coordinates of the 3D Gaussian \\
\textbf{$\boldsymbol{s}$} & Spatial extent occupied by the 3D Gaussian \\
\rowcolor{gray!25} \textbf{$\boldsymbol{r}$} & Orientation of the 3D Gaussian \\
\textbf{$\boldsymbol{h}$} & Colors that vary with the viewing direction \\
\rowcolor{gray!25} \textbf{$d$} & Degree of the Spherical Harmonics \\
\textbf{$\boldsymbol{\Sigma}$} & Covariance matrix of the 3D Gaussian \\
\rowcolor{gray!25} \textbf{$\boldsymbol{R}$} & Rotation matrix of the 3D Gaussian \\
\textbf{$\boldsymbol{S}$} & Scale matrix of the 3D Gaussian \\
\rowcolor{gray!25} \textbf{$\sigma$} & Opacity value of each 3D Gaussian \\
\textbf{$M$} & Bit width of the 3D Gaussian attributes \\
\rowcolor{gray!25} \textbf{$G$} & Individual 3D Gaussian \\
\textbf{$\mathcal{G}$} & Set of 3D Gaussians at a specific camera pose \\
\rowcolor{gray!25} \textbf{$C$} & View-dependent RGB color of the Gaussian \\
\\
\multicolumn{2}{l}{\textbf{2. Rendering and Projection}} \\
\rowcolor{gray!25} \textbf{$\boldsymbol{W}$} & Coordinate Transformation matrix from world coordinates to camera coordinates \\
\textbf{$T$} & Tile size \\
\rowcolor{gray!25} \textbf{$\boldsymbol{p}$} & Pixel in 2D space \\
\textbf{$\boldsymbol{\mu}'$} & Projected center of the 3D Gaussian in 2D screen coordinates system \\
\rowcolor{gray!25} \textbf{$\boldsymbol{\Sigma}'$} & Projected 2D covariance matrix of the 3D Gaussian in screen coordinates \\
\textbf{$\mathcal{N}$} & Total number of Gaussians projected onto the pixel \\
\rowcolor{gray!25} \textbf{$\alpha(\boldsymbol{p})$} & Alpha blending value at pixel $p$ \\
\textbf{$g(\boldsymbol{p})$} & Pixel value of the projected 2D Gaussian in screen coordinates \\
\rowcolor{gray!25} \textbf{$c(\boldsymbol{p})$} & Predicted pixel color value \\
\textbf{$I_{pred}$} & Rendered 2D image \\
\rowcolor{gray!25} \textbf{$I_{gt}$} & GT image \\
\textbf{$\phi$} & Transformation matrix that defines the camera's viewpoint and orientation \\
\\
\multicolumn{2}{l}{\textbf{3. Static Scene Modeling}} \\
\rowcolor{gray!25} {$\boldsymbol{\mathcal{A}}$} & Gaussian attributes \\
{$\mathcal{M}_n$} & Binary mask for n-th 3D Gaussian \\
\rowcolor{gray!25} {$m_n$} & Learnable mask parameter for n-th 3D Gaussian \\
{$\epsilon$} & Threshold value for masking \\
\rowcolor{gray!25} {$\operatorname{sg}(\cdot)$} & Stop Gradient operator \\
{$\mathds{1}[\cdot]$} & Indicator function \\
\rowcolor{gray!25} {$\operatorname{Sig}(\cdot)$} & Sigmoid activation function \\
{$\hat{\boldsymbol{s}}$} & Masked scale of 3D Gaussian \\
\rowcolor{gray!25} {$\hat{\sigma}$} & Masked opacity of 3D Gaussian \\
{$L_m$} & Masking loss \\
\rowcolor{gray!25} {$GS_j$} & Global Significance score for j-th Gaussian \\
{$\gamma(\boldsymbol{\Sigma}_j)$} & Normalized volume of j-th Gaussian \\
\rowcolor{gray!25} {$W_{i,p}$} & Influence of i-th Gaussian at pixel p \\
{$W_i$} & Total influence of i-th Gaussian on entire scene \\
\rowcolor{gray!25} {$\alpha_i$} & Alpha blending value of i-th Gaussian \\
{$\mathcal{T}_i$} & Transmittance value up to the i-th 3D Gaussian \\
\rowcolor{gray!25} {$\boldsymbol{\mathcal{N}_i^K}$} & Set of K-nearest neighbors of i-th 3D Gaussian \\
{$\boldsymbol{T}_i$} & Appearance feature vector of i-th 3D Gaussian \\
\rowcolor{gray!25} {$\mathcal{L}_{\texttt{distill}}$} & Knowledge distillation loss \\
{$\boldsymbol{C}_{\texttt{teacher}}$} & Rendered pixel values from teacher model \\
\rowcolor{gray!25} {$\boldsymbol{C}_{\texttt{student}}$} & Rendered pixel values from student model \\
{$S(p)$} & Sensitivity of parameter p \\
\rowcolor{gray!25} {$E_j$} & Total energy for j-th image \\
{$p_k$} & Probability of k-th codebook vector \\
\rowcolor{gray!25} {$r_{i,k}^{(s)}$} & Rate loss when k-th codeword is selected for the scale of i-th 3D Gaussian \\
{$d_{i,k}^{(s)}$} & Distortion loss when k-th codeword is selected for the scale of i-th 3D Gaussian \\
\rowcolor{gray!25} {$\texttt{\textbf{CB}}^{(\boldsymbol{s})}[k]$} & k-th codeword in scale codebook \\
{$\boldsymbol{P}$} & Point clouds from COLMAP \\
\rowcolor{gray!25} {$\boldsymbol{V}$} & Voxelized scene \\
{$\boldsymbol{f}_v$} & Local context features of anchor \\
\rowcolor{gray!25} {$\boldsymbol{l}_v$} & Scaling factor of anchor \\
{$\boldsymbol{O}_v$} & Learnable offsets of anchor \\
\rowcolor{gray!25} {$\hat{\boldsymbol{f}}_v$} & View-dependent features \\
{$\delta_{vc}$} & Relative distance between camera and anchor \\
\rowcolor{gray!25} {$\boldsymbol{d}_{vc}$} & View direction vector \\
{$F_{\boldsymbol{a}}$} & MLP for deriving attributes \\
\rowcolor{gray!25} {$D$} & MLP decoder function \\
{$\hat{\boldsymbol{q}}$} & Latent vector of the quantized attribute \\
\rowcolor{gray!25} {$\bar{\boldsymbol{q}}$} & Rounded latent vector \\
{$\boldsymbol{f}_h$} & Hash feature \\
\rowcolor{gray!25} {$q_i$} & Quantization step size for the i-th anchor \\
{$\hat{\boldsymbol{f}}_a^i$} & Quantized i-th anchor attribute \\
\rowcolor{gray!25} {$\phi_{\boldsymbol{\mu}_a^i, \boldsymbol{\sigma}_a^i}$} & Gaussian distribution for i-th quantized anchor attributes \\
{$\Phi_{\boldsymbol{\mu}_a^i, \boldsymbol{\sigma}_a^i}$} & Cumulative distribution function for i-th anchor attributes \\
\rowcolor{gray!25} {$\texttt{MLP}_c$} & MLP for context modeling \\
\\
\multicolumn{2}{l}{\textbf{4. Optimization and Training}} \\
\rowcolor{gray!25} \textbf{$x$} & Arbitrary point in world coordinates \\
\textbf{$I_{gt}$} & Ground truth image \\
\rowcolor{gray!25} \textbf{$L_1(\cdot)$} & Pixel-wise mean absolute error \\
\textbf{$\lambda$} & Weighting factor controlling relative contribution of $L_1$ and SSIM loss \\
\rowcolor{gray!25} \textbf{$\tau_{pos}$} & Threshold for view-space position gradients of a Gaussian \\
\textbf{$\epsilon_{\sigma}$} & Opacity threshold for the Gaussian \\
\\
\multicolumn{2}{l}{\textbf{5. Dynamic Scene Modeling}} \\
\rowcolor{gray!25} $\Delta\mathbf{x}$ & Offset for position produced by time-conditioned MLP \\
$\Delta\mathbf{r}$& Offset for rotation produced by time-conditioned MLP \\
\rowcolor{gray!25} $\Delta\mathbf{s}$ & Offset for scale produced by time-conditioned MLP \\
$\mathbf{x}_c$ & Position of canonical gaussian \\
\rowcolor{gray!25} $F_{\theta}$ & Time-conditioned MLP \\
$E_\phi$ & HexPlane encoder \\
\rowcolor{gray!25} \textbf{f} & Latent feature \\
$D_\psi$ & Lightweight MLP decoder network \\
\rowcolor{gray!25} $\boldsymbol{\mu}_{4D}$ & 4D position \\
$\boldsymbol{\Sigma}_{4D}$ & 4D covariance matrix \\
\rowcolor{gray!25} $\mathbf{h}_{4D}$ & Spherical harmonics of Fourer coefficients of 4D coordinates \\
\textbf{$\boldsymbol{q}$} & Quaternion representing orientation of 3D Gaussian over time \\
\rowcolor{gray!25} \textbf{$t$} & Time variable for modeling dynamic scenes \\
\bottomrule
\caption{\textbf{Notation and symbols used throughout this survey}. The table organizes mathematical notation into five categories: basic elements of 3D Gaussians, rendering and projection, static scene modeling, optimization and training, and dynamic scene modeling.}
\end{longtable}